\documentclass[lettersize,journal]{IEEEtran}

\usepackage{amsmath,amsfonts}
\usepackage{pifont}
\usepackage{mathtools}
\usepackage{algorithmic}
\usepackage{algorithm}
\usepackage{array}
\usepackage[caption=false,font=normalsize,labelfont=sf,textfont=sf]{subfig}
\usepackage{textcomp}
\usepackage{stfloats}
\usepackage{url}
\usepackage{verbatim}
\usepackage{graphicx}
\usepackage{cite}

\usepackage{booktabs}
\usepackage{nicefrac}
\usepackage{microtype}
\usepackage{xcolor}
\usepackage{graphicx}
\usepackage{enumitem}
\usepackage{color}
\usepackage{overpic}
\usepackage{pifont}
\usepackage{multirow}
\usepackage{bm}
\usepackage{mathrsfs}
\usepackage[normalem]{ulem}

\usepackage[colorlinks,
            linkcolor=red,
            anchorcolor=blue,
            citecolor=green,
            ]{hyperref}

\usepackage{amsmath,amsfonts,bm}

\def\1{\bm{1}}

\DeclareMathAlphabet{\mathsfit}{\encodingdefault}{\sfdefault}{m}{sl}
\SetMathAlphabet{\mathsfit}{bold}{\encodingdefault}{\sfdefault}{bx}{n}

\definecolor{mycyan}{RGB}{212, 239, 251}
\usepackage{colortbl}
\usepackage{xcolor}
\definecolor{mygray}{gray}{.9}
\definecolor{goldenrod}{RGB}{245,245,220}
\newlength\savewidth\newcommand\shline{\noalign{\global\savewidth\arrayrulewidth\global\arrayrulewidth 1pt}\hline\noalign{\global\arrayrulewidth\savewidth}}
\newcolumntype{a}{>{\columncolor{mygray}}c}
\usepackage{fontawesome5}
\usepackage{booktabs}
\usepackage{makecell}
\usepackage{fontawesome5}
\usepackage{lipsum}
\usepackage{comment}
\usepackage{multirow}
\usepackage{wrapfig}
\usepackage{algorithm}
\usepackage{algorithmic}
\usepackage{xfrac}  

\usepackage{graphicx}
\usepackage{color}

\usepackage[english]{babel}
\definecolor{darkgreen}{rgb}{0,0.7,0}

\newcommand{\green}[1]{{\color{darkgreen}{#1}}}
\definecolor{mygraytext}{gray}{.75}

\def\eg{\emph{e.g.}} 
\def\ie{\emph{i.e.}}

\def\etal{\emph{et al.}}

\newcommand{\as}[1]{\renewcommand{\arraystretch}{#1}}
\newcommand{\customstrut}[1]{\rule{0pt}{#1}\rule[-1ex]{0pt}{0pt}} 

\begin{document}

\title{Expandable Residual Approximation for \\ Knowledge Distillation}

\author{{Zhaoyi Yan, Binghui Chen, Yunfan Liu, Qixiang Ye}
\thanks{Z. Yan is with the School of Computer Science and Technology, Harbin Institute of Technology, Harbin, China (e-mail: yanzhaoyi@outlook.com).}
\thanks{B. Chen is with Beijing University of Posts and Telecommunications, Beijing, China (e-mail: chenbinghui@bupt.cn).}
\thanks{Y. Liu and Q. Ye are with the School of Electronics, Electrical and Communication Engineering, University of Chinese Academy of Sciences, Beijing, China (e-mail: liuyunfan@ucas.ac.cn, qxye@ucas.ac.cn).}
\thanks{Corresponding author: Yunfan Liu.}
}

\markboth{IEEE TRANSACTIONS ON Neural Networks and Learning Systems, Manuscript for Review}%
{Shell \MakeLowercase{\textit{et al.}}: A Sample Article Using IEEEtran.cls for IEEE Journals}

\maketitle

\begin{abstract}
Knowledge distillation (KD) aims to transfer knowledge from a large-scale teacher model to a lightweight one, significantly reducing computational and storage requirements.
However, the inherent learning capacity gap between the teacher and student often hinders the sufficient transfer of knowledge, motivating numerous studies to address this challenge.
Inspired by the progressive approximation principle in the Stone-Weierstrass theorem, we propose Expandable Residual Approximation (ERA), a novel KD method that decomposes the approximation of \textit{residual knowledge} into multiple steps, reducing the difficulty of mimicking teacher's representation through a divide-and-conquer approach.
Specifically, ERA employs a Multi-Branched Residual Network (MBRNet) to implement this residual knowledge decomposition. Additionally, a Teacher Weight Integration (TWI) strategy is introduced to mitigate the capacity disparity by reusing the teacher's head weights. 
Extensive experiments show that ERA improves the Top-1 accuracy on ImageNet classification benchmark by 1.41\% and the AP on the MS COCO object detection benchmark by 1.40, as well as achieving leading performance across computer vision tasks.
Codes and models are available at \url{https://github.com/Zhaoyi-Yan/ERA}.
\end{abstract}

\begin{IEEEkeywords}
Knowledge Distillation, Residual Learning, Image Classification, Visual Object Detection, Semantic Segmentation
\end{IEEEkeywords}

\section{Introduction}
\label{sec:intro}

\IEEEPARstart{L}{arge}-scale deep models have achieved remarkable success across a wide range of vision tasks~\cite{Simonyan2015vgg,he2016deep,Huang2017densenet,liu2022convnet,Alexey2021ViT,liu2021swin}. 
However, their computational complexity and substantial storage requirements pose significant challenges for deployment on resource-constrained devices. 
To conquer this issue, Hinton~\etal~\cite{hinton2015distilling} introduced Knowledge Distillation (KD), a technique that compresses a high-capacity yet resource-intensive teacher model to a smaller yet more efficient student model.
Due to its simplicity and effectiveness, KD has been widely adopted~\cite{chen2017learning,liu2019structured,deit}, and extensive research has been conducted to advance KD through improved knowledge representations~\cite{hinton2015distilling,ba2014deep,chen2020online,yuan2020revisiting,yang2020knowledge,payingICLR2017,tung2019similarity,heo2019knowledge,chen2021distilling,l2tICML2019,darkrankAAAI2018,tung2019similarity,park2019relational,peng2019correlation,sstaICML2022,giftCVPR2017,lee2018self,zhang2018better,lee2020graph,passalis2020heterogeneous,sstaICML2022,joshi2024GNN,Ding2024dual}, transfer strategies~\cite{wang2018kdgan,shen2019meal,wang2018adversarial,chen2015net2net,gou2024collaborative}, and distillation frameworks~\cite{vongkulbhisal2019unifying,yuan2021reinforced,mirzadeh2020improved,gao2021residual}.

\begin{figure*}[!t]
\centering
\begin{overpic}[scale=0.7]{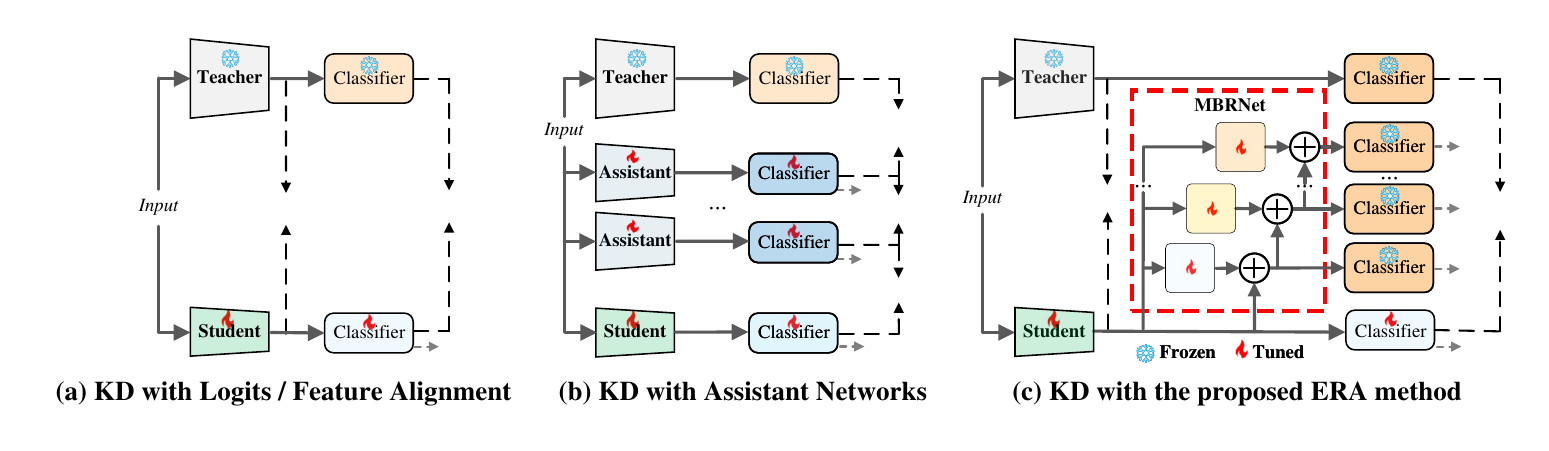}

\put(17.7, 15.5){\scalebox{.60}{$\mathcal{L}_2$}}
\put(27.5, 15.5){\scalebox{.60}{$\mathcal{L}_{KL}$}}
\put(27.0, 5.5){\scalebox{.60}{$\mathcal{L}_{cls}$}}

\put(57.0, 20.5){\scalebox{.60}{$\mathcal{L}_{KL}$}}
\put(57.0, 15.3){\scalebox{.60}{$\mathcal{L}_{KL}$}}
\put(57.0, 10.0){\scalebox{.60}{$\mathcal{L}_{KL}$}}

\put(54.0, 5.5){\scalebox{.60}{$\mathcal{L}_{cls}$}}
\put(54.0, 11.0){\scalebox{.60}{$\mathcal{L}_{cls}$}}
\put(54.0, 15.5){\scalebox{.60}{$\mathcal{L}_{cls}$}}

\put(70.5, 16.0){\scalebox{.60}{$\mathcal{L}_2$}}
\put(95.5, 15.3){\scalebox{.60}{$\mathcal{L}_{KL}$}}

\put(93.5, 5.5){\scalebox{.60}{$\mathcal{L}_{cls}$}}
\put(93.5, 10.5){\scalebox{.60}{$\mathcal{L}_{cls}$}}
\put(93.5, 14.5){\scalebox{.60}{$\mathcal{L}_{cls}$}}
\put(93.5, 18.0){\scalebox{.60}{$\mathcal{L}_{cls}$}}

\end{overpic}
\caption{Framework comparison between (a) one-step knowledge distillation (KD) with logits and feature alignment~\cite{hinton2015distilling,romero2014fitnets,simKDCVPR2022}, (b) KD with assistant networks~\cite{mirzadeh2020improved}, where assistant networks are sequentially constructed from the previous teacher (or assistant network), making the process parameter- and computation-intensive, and (c) the proposed ERA approach. The ERA framework models the capability gap between the teacher and student (\ie, the `residual knowledge') progressively using a multi-branched residual network architecture, MBRNet.
}
\label{fig:compare}
\end{figure*}
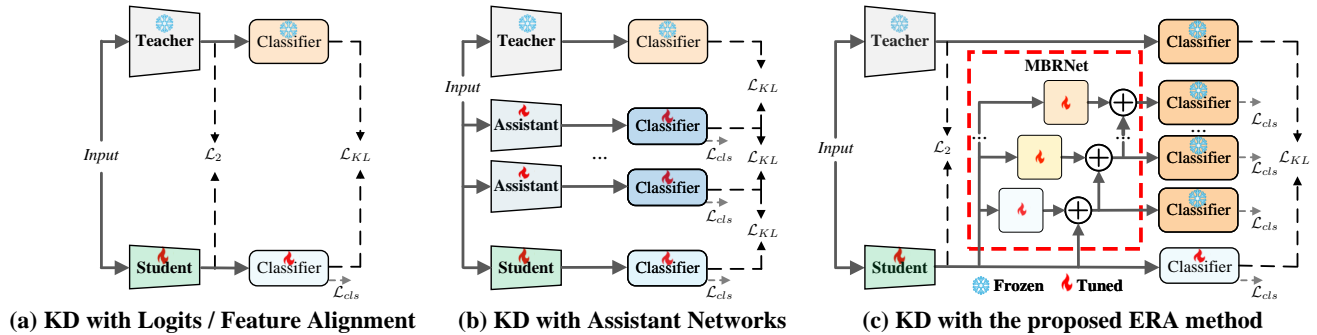

In vanilla KD~\cite{hinton2015distilling}, the student model is trained to mimic the prediction of the teacher model by aligning their output logits under the same input, a process commonly referred to as \textit{logits-based alignment}.
However, the inherent disparity in learning capacity prevents the student from fully capturing the knowledge from the teacher, thereby limiting its overall effectiveness.
To address this limitation, FitNets~\cite{romero2014fitnets} proposes to use intermediate feature maps from the teacher model as auxiliary knowledge to guide the distillation process, Fig.~\ref{fig:compare}(a).
This approach, referred to as \textit{feature-based alignment}, has inspired significant efforts to explore advanced knowledge representations beyond plain feature maps~\cite{huang2017like,zagoruyko2022paying} and similarity patterns~\cite{passalis2018learning,tung2019similarity}, and to design improved feature alignment losses~\cite{ahn2019variational, park2019relational, passalis2020probabilistic}.
However, as noted by~\cite{huang2022knowledge, mirzadeh2020improved}, strictly enforcing feature alignment can conflict with the task-specific objective.
Moreover, a large capability gap between teacher and student models aggregates the overfitting risk when strict output or feature alignment is enforced.

To bridge the capacity gap between teacher and student models, recent studies~\cite{mirzadeh2020improved, son2021densely, sstaICML2022} propose incorporating auxiliary networks or additional distillation steps into the conventional framework. 
TAKD~\cite{mirzadeh2020improved} introduces teacher assistant (TA) networks of intermediate sizes, progressively transferring knowledge to the student through a series of TAs with decreasing capacities, Fig.~\ref{fig:compare} (b). 
This process requires iterative distillation over multiple rounds, which can be computationally expensive and time-consuming.
%
In contrast, Gao~\etal~\cite{gao2021residual} propose a single-step distillation method using one TA to capture the residual error between teacher and student representations, reformulating the objective into two complementary lightweight models.
Due to the significant capacity gap, however, such `residual knowledge'  exhibits complex patterns, making it challenging to model effectively with a lightweight assistant network. 
To date, strategies for addressing this challenge remain unexplored.

In this study, we build upon the concept of residual learning and propose Expandable Residual Approximation (ERA), a novel knowledge distillation (KD) approach that \textbf{decomposes the residual knowledge through multi-step distillation}. 
Specifically, ERA employs a Multi-Branched Residual Network (MBRNet) to iteratively approximate the capability gap between teacher and student models, reducing the complexity of feature alignment and mitigating the risk of overfitting.
As illustrated in Fig.~\ref{fig:compare} (c), MBRNet takes the feature representation generated by the student model as input and progressively approximates the teacher model's predictions by accumulating outputs from its branches.
Notably, the incorporation of MBRNet enables ERA to support three distinct inference modes, namely generating student logits ($S$-mode), teacher logits ($T$-mode), or a combination of both ($ST$-mode), offering greater flexibility and adaptability in various scenarios (see Fig.~\ref{fig:diagram}).
To further enhance ERA, we propose a Teacher Weight Integration (TWI) strategy, which reuses the pre-trained weights of the teacher classifier to compute the logits for each MBRNet branch. 
By incorporating the capability-specific information embedded in the teacher's discriminative classifier~\cite{simKDCVPR2022}, TWI alleviates the capacity gap and improves the task-specific performance of the mimicked teacher features.
%

The contributions of this study are summarized as follows:
\begin{enumerate}[label=\textbullet]

    \item We propose Expandable Residual Approximation (ERA), a novel knowledge distillation (KD) method that leverages a Multi-Branched Residual Network (MBRNet) to decompose the residual knowledge and progressively bridge the gap in model capacity.

    \item We propose a Teacher Weight Integration (TWI) strategy, which can leverage the mimicked teacher features to boost the task-specific performance of both the student backbone and MBRNet, enhancing the overall model's efficacy.
    
    \item We validate that ERA consistently outperforms state-of-the-art methods, highlighting its effectiveness across inference modes.
   
\end{enumerate}

\section{Related Work}\label{sec:related_work}

\subsection{Knowledge Representation}
The goal of KD is to compress the knowledge of a powerful, large-scale teacher model to a lightweight and computationally efficient student model.
Therefore, representing the knowledge to be transferred is central to the process, and it has been the subject of extensive research.
Hinton~\etal~\cite{hinton2015distilling}, in their seminal work, introduced the concept of knowledge distillation, where the softened logits distribution (soft targets) generated by a teacher model encodes informative `dark knowledge'. 
This knowledge is transferred to a student model by minimizing the Kullback–Leibler (KL) divergence between their output distributions. 
%
Subsequent studies~\cite{ba2014deep,chen2020online,yuan2020revisiting} demonstrated that soft targets effectively capture the relationships between classes, serving as a regularizer to guide the distillation process. 
Furthermore, NKD~\cite{nkd} refined the distillation objective by redefining its connection to cross-entropy, while SRRL~\cite{yang2020knowledge} utilized the teacher model’s projection matrix to enhance the student’s representation capability.

Due to the significant gap in model capacity, the student model struggles to approximate the predictions generated by deeper layers of the teacher, posing challenges for knowledge transfer. 
To address this issue, Romero~\etal~\cite{romero2014fitnets} propose aligning the feature maps of the teacher and student models, as both the final output and intermediate layers of deep models encode rich visual representations.
This idea inspired subsequent studies\cite{payingICLR2017,tung2019similarity,heo2019knowledge}, which explored extensions of the vanilla feature maps to improve knowledge transfer.
Alternatively, to identify better feature associations for distillation, other works~\cite{chen2021distilling,l2tICML2019} investigated the impact of connection paths across intermediate layers between teacher and student models.
Chen~\etal~\cite{simKDCVPR2022} systematically compared KD schemes based on logits and feature alignment, analyzing differences in terms of formalization and starting point of the gradient flow. 

In addition to logits and feature maps, the relationships between feature maps and data samples, can also serve as a form of knowledge representation~\cite{darkrankAAAI2018,tung2019similarity,park2019relational,peng2019correlation,sstaICML2022}. 
Yim~\etal~\cite{giftCVPR2017} utilized the Gram matrix to capture relationships between feature map pairs, while Lee~\etal~\cite{lee2018self} distilled feature correlations via singular value decomposition.
Furthermore, the graph of knowledge was adopted in subsequent works~\cite{zhang2018better,lee2020graph,passalis2020heterogeneous,Ding2024dual} to model complex relationships between logits and feature maps.
Beyond logits and activations, the relationships between data samples also encode knowledge worthy of transfer.
Liu~\etal~\cite{liu2019knowledge} introduced the instance relation graph, which leveraged both the features and relationships of samples to represent knowledge. 
Chen~\etal~\cite{chen2020learning} proposed relational KD, where the student learns to preserve feature similarity between samples as estimated by the teacher.
SSTA~\cite{sstaICML2022} integrated supervision from both a supervised and a self-supervised teacher, using the latter as an assistant to enhance the student's learning process.

\subsection{Knowledge Transfer Strategy}
%
Although KD can be achieved by directly aligning various forms of  representation between student and teacher, more advanced transfer strategies were explored to further enhance the performance of distilled models.
Inspired by the success of Generative Adversarial Networks (GANs) in image generation~\cite{goodfellow2014generative,radford2016unsupervised,karras2017progressive,karras2019style,karras2020analyzing,karras2021alias}, the idea of adversarial learning was also introduced to the field of KD.
Recent studies leveraged pre-trained GANs to generate synthetic samples~\cite{chen2019data,micaelli2019zero} or augment datasets~\cite{wang2018kdgan,shen2019meal}, while others adopted adversarial objectives to train the student model to mimic the teacher’s logits and features, confusing a discriminative network~\cite{wang2018adversarial}.
The pre-training technique offers an informative parameter initialization strategy for the student model, where selected layers are pre-trained to mimic the corresponding teacher layers~\cite{romero2014fitnets,chen2015net2net}.
Multi-teacher distillation is a common setting in KD, where multiple teacher models with distinct knowledge are used to train a single student model~\cite{vongkulbhisal2019unifying,yuan2021reinforced,gou2024collaborative}. 
In contrast, some studies~~\cite{mirzadeh2020improved,gao2021residual} proposed incorporating assistant models to bridge the capacity gap between teacher and student models, as discussed in the following subsection.

\subsection{KD with Assistant Networks}
Despite advancements in the field of KD, a notable challenge persists: as the size gap between teacher and student models increases, the performance of the student model becomes inconsistent, highlighting the difficulty of knowledge transfer.
%
To address this limitation, several approaches have been proposed. 
TAKD~\cite{mirzadeh2020improved} introduced intermediate-sized networks, known as teacher assistants (TAs), to bridge the capacity gap between teachers and students.
SimKD~\cite{simKDCVPR2022} simplified this process by directly integrating the discriminative classifier of a pre-trained teacher model into the student backbone, which can also be considered as a kind of assistant model.
TinyBERT~\cite{tinybert}, a pioneering method in natural language processing, adopted a two-stage knowledge distillation framework. 
ResErrKD~\cite{gao2021residual} improved the knowledge transfer process by introducing an assistant model that learns the residual error between the teacher and student models.
ResKD~\cite{li2021reskd} further extended this residual-guidance mechanism into an iterative framework, enabling users to flexibly balance accuracy and computational cost through repeated refinement.
%
Recently, Generic-to-Specific Distillation (G2SD)~\cite{huang2023generic} improved knowledge transfer by first aligning the student's decoder with the teacher's representations (generic distillation) and then ensuring prediction consistency for task-specific insights (specific distillation).
Despite these advancements, the performance gap between complex teacher models and their distilled student counterparts remains. 
To conquer this issue, we propose a novel method ERA to bridge this gap by progressively enhancing the student's ability to approximate the teacher's predictions.

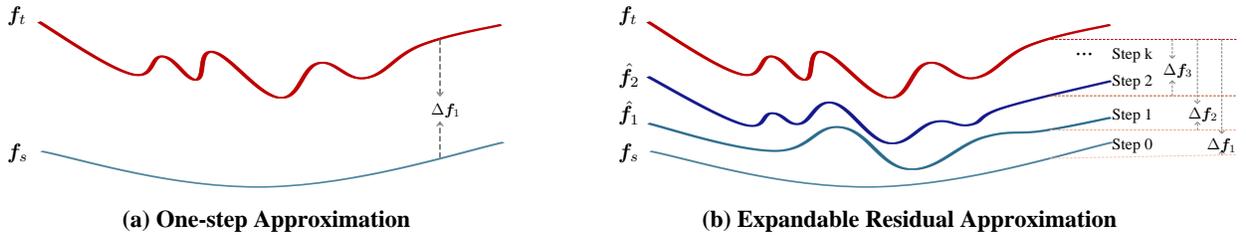
\begin{figure*}[t]
\centering
\begin{overpic}[scale=0.5]{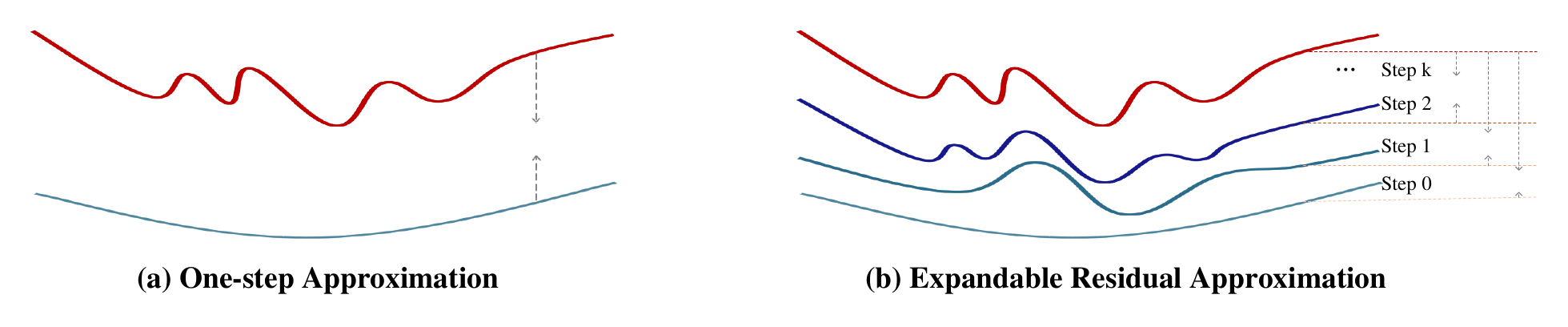} 

\put(-0.5, 8.0){\scalebox{.80}{$\boldsymbol{f}_s$}}
\put(-0.5, 19.0){\scalebox{.80}{$\boldsymbol{f}_t$}}
\put(33.5, 11.5){\scalebox{.60}{$\Delta\boldsymbol{f}_1$}}

\put(48.5, 19.0){\scalebox{0.80}{$\boldsymbol{f}_t$}}
\put(48.5, 14.0){\scalebox{0.80}{$\hat{\boldsymbol{f}}_2$}}
\put(48.5, 11.0){\scalebox{0.80}{$\hat{\boldsymbol{f}}_1$}}
\put(48.5, 8.0){\scalebox{0.80}{$\boldsymbol{f}_s$}}

\put(95.7, 8.7){\scalebox{.60}{$\Delta\boldsymbol{f}_1$}}
\put(94.2, 11.2){\scalebox{.60}{$\Delta\boldsymbol{f}_2$}}
\put(92.2, 14.7){\scalebox{.60}{$\Delta\boldsymbol{f}_3$}}

\end{overpic}
\scriptsize
\caption{Illustration of (a) the one-step approximation used in conventional KD methods~\cite{hinton2015distilling,romero2014fitnets}, and (b) the multi-step approximation introduced by the proposed ERA method. ERA decomposes the model capacity gap between the teacher ($\boldsymbol{f}_t$) and the student ($\boldsymbol{f}_s$) into smaller sub-problems, allowing for easier and more effective approximation at each step.
}
\label{fig:illustration}
\end{figure*}

\section{Preliminary}\label{sec:preliminary}

In this section, we provide a brief review of knowledge distillation (KD) based on logits and feature alignment.
As shown in Fig.~\ref{fig:compare}, a classification network consists of a feature extractor and a softmax classifier~\cite{he2016deep,howard2017mobilenets,zhang2018shufflenet}, which are optimized end-to-end via gradient back-propagation.
Unless otherwise specified, the terms `teacher model' and `student model' refer to the feature extractor (encoder) network.

\subsection{KD with Logit Alignment}
Mathematically, given an input image $\boldsymbol{x}$, its representation (\ie, feature vector) $\boldsymbol{f}_s \in \mathbb{R}^{C_s}$, computed by the student model $\mathcal{F}_s$, can be expressed as
\begin{equation}
    \boldsymbol{f}_s=\mathcal{F}_s(\boldsymbol{x};\boldsymbol{\theta}_s),
\end{equation}
where $\boldsymbol{\theta}_s$ is the parameter of the student model.
Subsequently, $\boldsymbol{f}_s$ is passed through a classifier network $\boldsymbol{h}_{s}$ with weights $\boldsymbol{W}_s \in \mathbb{R}^{M \times C_s}$, where $M$ is the total number of categories.
The output logits $\boldsymbol{g}_s$ and class probabilities $\boldsymbol{p}_s$ are given by
\begin{equation}
    \boldsymbol{g}_s=\boldsymbol{W}_s\boldsymbol{f}_s
\end{equation}
and 
\begin{equation}
    \boldsymbol{p}_s=\sigma(\boldsymbol{g}_s/T),
\end{equation}
respectively, where $\sigma(\cdot)$ denotes the softmax function and $T$ is a temperature parameter.
For the teacher model $\mathcal{F}_t$, the corresponding feature vector $\boldsymbol{f}_t$, logits $\boldsymbol{g}_t$, and class probabilities $\boldsymbol{p}_t$ are computed in a similar manner.

In the conventional KD framework based on logits alignment, the objective function mainly consists of two components: a cross-entropy loss $\mathcal{L}_{\text{CE}}$ for input classification and a distillation loss $\mathcal{L}_{\text{KL}}$ that measures the discrepancy between the predicted logits of teacher and student via Kullback–Leibler (KL) divergence.
Specifically, the overall loss function for logits distillation (denoted as $\mathcal{L}_{\text{LD}}$) can be expressed as
\begin{equation}\label{eq:KD_logits}
    \mathcal{L}_{\text{LD}}=\alpha\mathcal{L}_{\text{CE}}(\boldsymbol{y}, \boldsymbol{p}_s) + \beta T^2\mathcal{L}_{\text{KL}}(\boldsymbol{p}_t, \boldsymbol{p}_s),
\end{equation}
where $\alpha$ and $\beta$ are hyper-parameters balancing the weight of the corresponding loss term, and $\boldsymbol{y}$ is the ground-truth label of the input image $\boldsymbol{x}$.

\subsection{KD with Feature Alignment}
In addition to aligning logits, KD methods based on feature alignment improve the student's representational capacity by mimicking the teacher model's intermediate feature maps.
Concretely, given the feature vector of the teacher model ($\boldsymbol{f}_t = \mathcal{F}_t(\boldsymbol{x};\boldsymbol{\theta}_t) \in \mathbb{R}^{C_t}$) and the student model ($\boldsymbol{f}_s = \mathcal{F}_s(\boldsymbol{x};\boldsymbol{\theta}_s) \in \mathbb{R}^{C_s}$), the objective function for feature distillation (denoted as $\mathcal{L}_{\text{FD}}$) can be formulated as follows
\begin{equation}
    \mathcal{L}_{\text{FD}}=\frac{1}{N}\sum_{n=1}^{N}\|\boldsymbol{f}_t^n - \mathcal{P}\boldsymbol{f}_s^n\|^2,
    \label{eqn:feat_dis}
\end{equation}
where $N$ is the number of training samples, $\|\cdot\|^2$ denotes the squared Euclidean norm, and $\mathcal{P} \in \mathbb{R}^{C_t \times C_s}$ maps the student's features to a channel space compatible with the teacher's.

\section{Methodology}\label{sec:method}

\begin{figure*}[!t]
\centering
\begin{overpic}[scale=0.6]{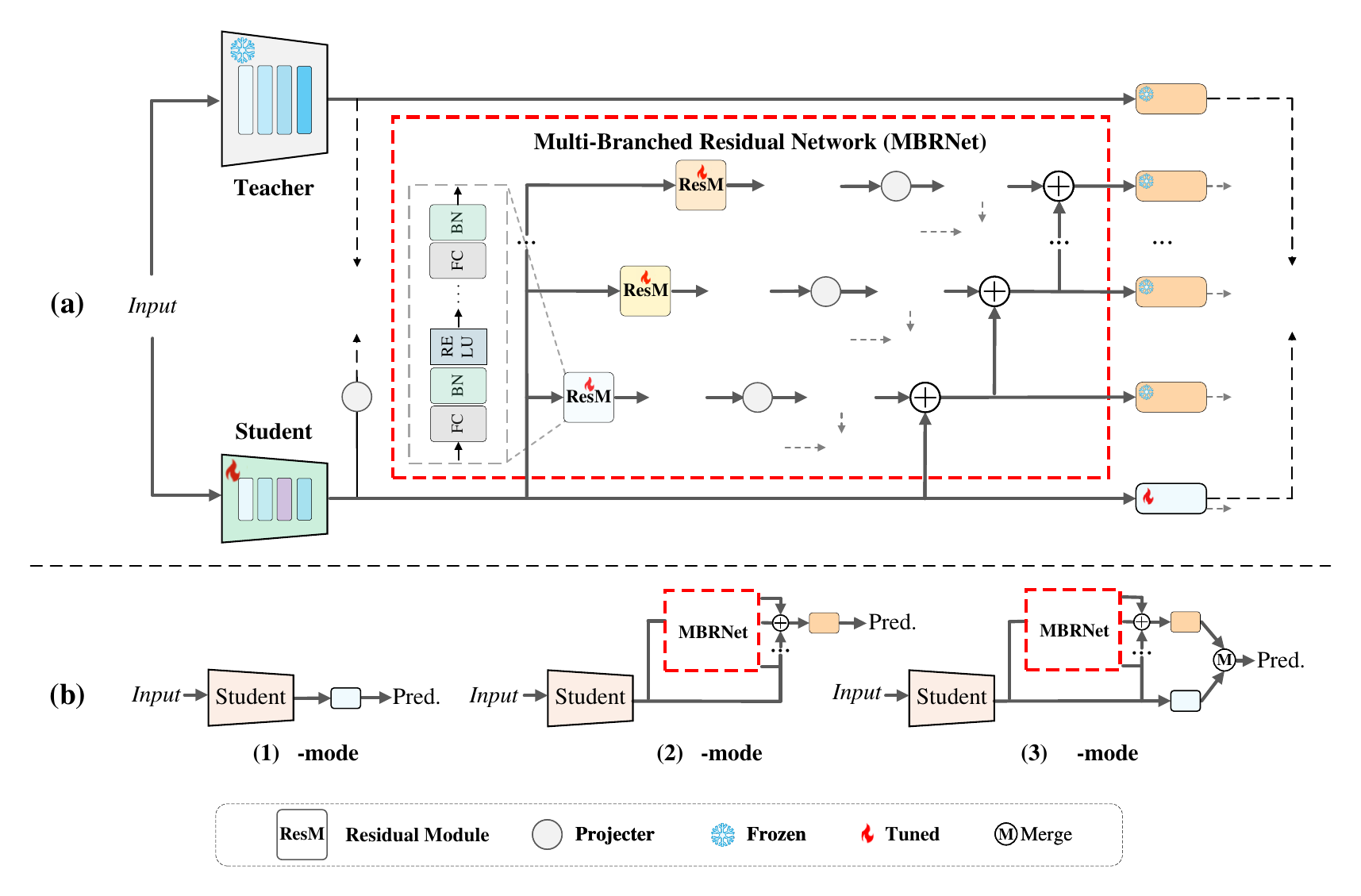} 

\put(24.5, 27.3){\scalebox{.70}{$\boldsymbol{f}_{s}$}}
\put(24.5, 57.1){\scalebox{.70}{$\boldsymbol{f}_{t}$}}
\put(24.5, 43.5){\scalebox{.70}{$\mathcal{L}_{FD_0}$}}

\put(25.4, 36.2){\scalebox{.60}{$\mathcal{P}_0$}}
\put(55.0, 36.3){\scalebox{.60}{$\mathcal{P}_1$}}
\put(60.0, 44.1){\scalebox{.60}{$\mathcal{P}_2$}}
\put(65.1, 51.8){\scalebox{.60}{$\mathcal{P}_K$}}

\put(85.6, 28.8){\scalebox{.80}{$\boldsymbol{h}_s$}}
\put(85.6, 36.3){\scalebox{.80}{$\boldsymbol{h}_t$}}
\put(85.6, 44.0){\scalebox{.80}{$\boldsymbol{h}_t$}}
\put(85.6, 51.7){\scalebox{.80}{$\boldsymbol{h}_t$}}
\put(85.6, 58.0){\scalebox{.80}{$\boldsymbol{h}_t$}}

\put(48.3, 36.2){\scalebox{.80}{$\Delta\hat{\boldsymbol{f}}_1$}}
\put(52.7, 44.0){\scalebox{.80}{$\Delta\hat{\boldsymbol{f}}_2$}}
\put(57.3, 51.7){\scalebox{.80}{$\Delta\hat{\boldsymbol{f}}_K$}}

\put(79.0, 34.5){\scalebox{.80}{$\hat{\boldsymbol{f}}_1$}}
\put(79.0, 42.5){\scalebox{.80}{$\hat{\boldsymbol{f}}_2$}}
\put(79.0, 50.3){\scalebox{.80}{$\hat{\boldsymbol{f}}_K$}}

\put(54.5, 32.5){\scalebox{.80}{$\Delta\boldsymbol{f}_1$}}
\put(59.4, 40.5){\scalebox{.80}{$\Delta\boldsymbol{f}_2$}}
\put(64.0, 48.5){\scalebox{.80}{$\Delta\boldsymbol{f}_K$}}

\put(60.7, 32.5){\scalebox{.80}{$\mathcal{L}_{FD_1}$}}
\put(65.6, 40.5){\scalebox{.80}{$\mathcal{L}_{FD_2}$}}
\put(70.7, 48.5){\scalebox{.80}{$\mathcal{L}_{FD_K}$}}

\put(60.0, 36.2){\scalebox{.60}{$\mathcal{P}_1\Delta\hat{\boldsymbol{f}}_1$}}
\put(65.0, 44.0){\scalebox{.60}{$\mathcal{P}_2\Delta\hat{\boldsymbol{f}}_2$}}
\put(69.5, 51.7){\scalebox{.60}{$\mathcal{P}_K\Delta\hat{\boldsymbol{f}}_K$}}

\put(89.5, 34.8){\scalebox{.70}{$\mathcal{L}_{cls_1}$}}
\put(89.5, 42.5){\scalebox{.70}{$\mathcal{L}_{cls_2}$}}
\put(89.5, 50.2){\scalebox{.70}{$\mathcal{L}_{cls_K}$}}

\put(93.8, 43.5){\scalebox{.70}{$\mathcal{L}_{KD}$}}
\put(89.5, 27.0){\scalebox{.70}{$\mathcal{L}_{CE}$}}

\put(20.7, 10.0){\scalebox{.70}{$\boldsymbol{S}$}}
\put(50.4, 10.0){\scalebox{.70}{$\boldsymbol{T}$}}
\put(77.0, 10.0){\scalebox{.70}{$\boldsymbol{ST}$}}

\end{overpic}
\scriptsize
\caption{Illustration of (a) the knowledge distillation framework based on the proposed Expandable Residual Approximation (ERA) approach, and (b) the inference modes enabled by different combinations of student model and trained MBRNet. The entire network is trained in a single stage, ensuring all residuals ($\mathcal{P}_k\Delta\hat{\boldsymbol{f}}_k$, $k=1,\dots,K$) are learned simultaneously.
}
\label{fig:diagram}
\end{figure*}
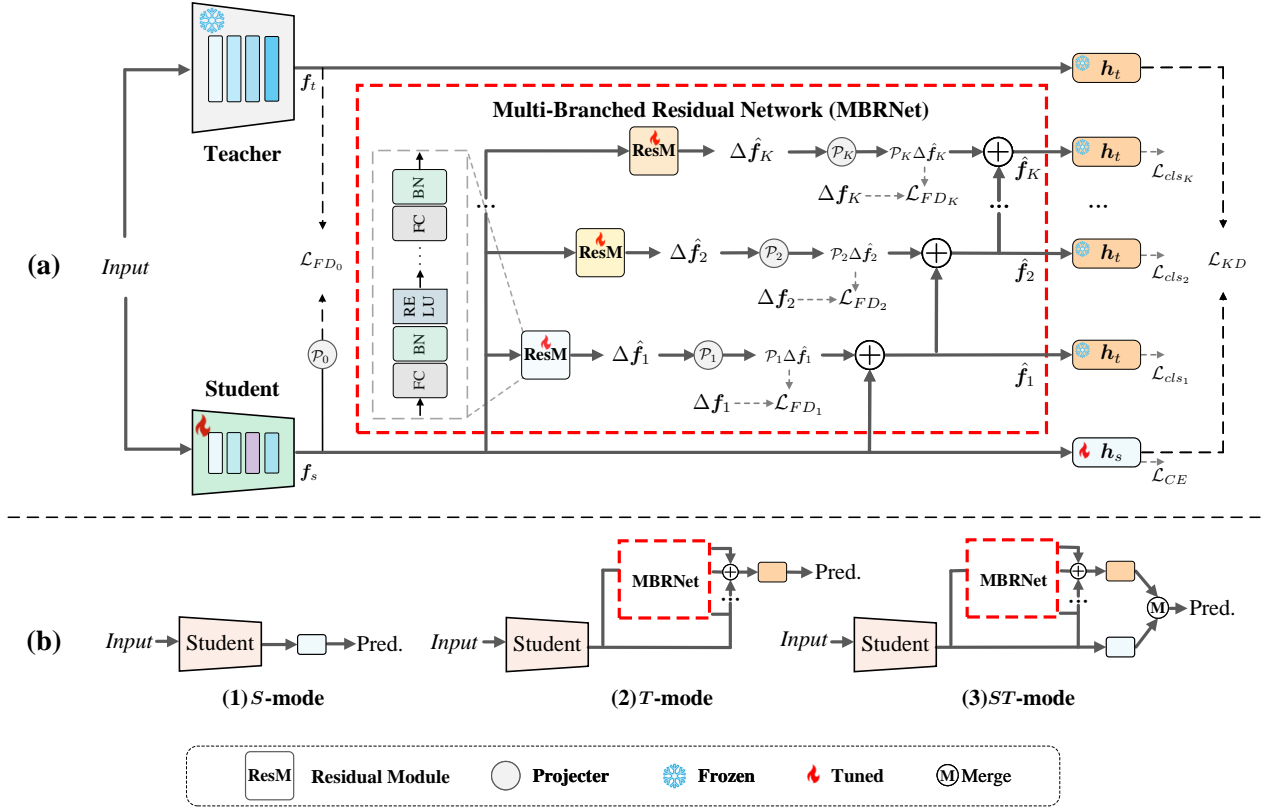

In this section, we first establish the theoretical foundation of the proposed ERA approach based on the Stone-Weierstrass theorem and then describe its implementation using a Multi-Branched Residual Network (MBRNet).
Additionally, we present the Teacher Weight Integration (TWI) strategy, which leverages the pre-trained weights of the teacher classifier to compute logits for the different branches of MBRNet.
Fig.~\ref{fig:diagram} illustrates the ERA framework and its inference modes, enabled by various configurations of the student model and MBRNet.

\subsection{Expandable Residual Approximation (ERA)}\label{sec:ERA}

In functional analysis, the Stone-Weierstrass theorem is a foundational result in function approximation. It asserts that any continuous function on a compact space can be uniformly approximated to arbitrary precision by polynomials or other function families satisfying specific algebraic and separation properties.
Specifically, for any given $\epsilon > 0$, there exists an integer $K > 0$ such that
\begin{align}\label{eq:SW_theorem}
    \|f-\sum_{i=0}^{K}{p_{i}g_{i}}\|<\epsilon,
\end{align}
where $f$ is the target continuous function, $p_i$ and $g_i$ are continuous functions chosen to approximate $f$, and $|\cdot|$ denotes a norm function measuring the error.

Inspired by the Stone-Weierstrass theorem, we propose the ERA method, which aims to approximate the teacher model by iteratively introducing auxiliary functions parameterized by deep neural networks (DNNs).
In this context, Eq.~\eqref{eq:SW_theorem} can be reformulated for the KD setting as
\begin{equation}\label{eq:SW_theorem_KD}
  \|\boldsymbol{f}_t - \sum_{i=0}^{K}\mathcal{P}_i \Delta\hat{\boldsymbol{f}}_i\|<\epsilon,
\end{equation}
where $\boldsymbol{f}_t$ denotes the feature maps of the teacher model, while $\mathcal{P}_j$ and $\Delta \hat{\boldsymbol{f}}_j$ are functions implemented via DNNs for progressive approximation.

Notably, if the output of the student network ($\boldsymbol{f}_s$) is treated as a zero-order approximation\footnote{We use $\boldsymbol{f}_s$ and $\Delta\hat{\boldsymbol{f}}_0$ interchangeably to refer to student network output.}, \ie, $\boldsymbol{f}_s \coloneqq \Delta\hat{\boldsymbol{f}}_0$, then Eq.~\eqref{eq:SW_theorem_KD} can be reformulated as
\begin{equation}\label{eq:SW_theorem_KD_reform}
  \|(\boldsymbol{f}_t - \mathcal{P}_0\boldsymbol{f}_s) - \sum_{i=1}^{K}\mathcal{P}_i \Delta\hat{\boldsymbol{f}}_i\|<\epsilon.
\end{equation}
Here, the overall objective can be regarded as approximating the residual knowledge, which is conceptually represented by $\boldsymbol{f}_t - \mathcal{P}_0\boldsymbol{f}_s$, by the output of $K$ DNNs.
As shown in Fig.~\ref{fig:illustration}, KD with feature alignment, Eq.~\eqref{eqn:feat_dis}, can be viewed as a simple one-step function approximation approach. 
In contrast, ERA employs a multi-step strategy to progressively reduce the gap between $\boldsymbol{f}_s$ and $\boldsymbol{f}_t$ over $K$ iterations, effectively breaking the problem into smaller, more manageable steps and reducing the difficulty of each individual approximation.

%


\subsection{Multi-Branched Residual Network (MBRNet)}\label{sec:MBRNet}

To compute the approximation functions $\{\Delta\hat{\boldsymbol{f}}_i\}_{i=1}^K$, we propose a Multi-Branched Residual Network (MBRNet) with $K$ branches, where each branch corresponds to one function.
As illustrated in Fig.~\ref{fig:diagram} (a), MBRNet employs a \textit{cascaded architecture} to achieve the multi-step approximation process of ERA.
Specifically, the $k$-th approximation of ERA, denoted as $\hat{\boldsymbol{f}_k}$ ($k \in \{1, \dots, K\}$), is computed by summing up the outputs of the first $k$ branches of MBRNet, starting from the student's prediction $\boldsymbol{f}_s$, as
\begin{equation}\label{eqn:delta_f}
     \hat{\boldsymbol{f}_k} = \boldsymbol{f}_s + \sum_{i=1}^k \mathcal{P}_i \Delta\hat{\boldsymbol{f}}_i,
\end{equation}
This expression can be further simplified to $\hat{\boldsymbol{f}_k} = \sum_{i=0}^k \mathcal{P}_i \Delta\hat{\boldsymbol{f}}_i$ by considering $\boldsymbol{f}_s$ as the initial approximation.

With the ultimate goal of mimicking the output of the teacher model $\boldsymbol{f}_t$, the training target for the $k$-th branch of MBRNet is to predict the residual knowledge $\Delta \boldsymbol{f}_k$ after $k-1$ steps of approximation, which can be expressed as
\begin{equation}\label{eqn:target_delta_f}
    \Delta \boldsymbol{f}_k = \boldsymbol{f}_t - \hat{\boldsymbol{f}}_{k-1} = \boldsymbol{f}_t - \sum_{i=0}^{k-1}\mathcal{P}_i \Delta \hat{\boldsymbol{f}}_i.
\end{equation}
Based on this, we can now derive the objective function for training MBRNet.
Similar to Eq.~\eqref{eqn:feat_dis}, the feature distillation loss at the $k$-th step ($k\in\{0,\dots,K\}$), denoted as $\mathcal{L}_{\text{FD}_k}$, is defined as
\begin{equation}
    \mathcal{L}_{\text{FD}_k} = \frac{1}{N}\sum_{n=1}^{N}\|\Delta\boldsymbol{f}_k^n - \mathcal{P}_{i}\Delta\hat{\boldsymbol{f}}_k^n\|^2,
    \label{eqn:feat_dis_branch}
\end{equation}
where $N$ denotes the total number of samples and $\boldsymbol{f}_k^n$ denotes the $n$-th sample of the student feature.
By optimizing all $K$ branches in MBRNet using Eq.~\eqref{eqn:feat_dis_branch}, the alignment between $\boldsymbol{f}_s$ and $\boldsymbol{f}_t$ is achieved progressively over multiple steps, rather than through a single-step approximation as described by Eq.~\eqref{eqn:feat_dis}.

In practical implementation, each branch in MBRNet (denoted as `ResM' in Fig.~\ref{fig:diagram}) consists of a stack of $m$ `Fully Connected (FC) $\rightarrow$ BatchNorm (BN) $\rightarrow$ ReLU' modules.
We omit the last ReLU layer to expand the range of output $\{\Delta \hat{\boldsymbol{f}_k}\}_{k=1}^K$, allowing their values to be positive or negative.
Moreover, the feature projection networks $\{\mathcal{P}_k\}_{k=0}^K$ are implemented using a single FC layer.
%
%

\subsection{Teacher Weight Integration}\label{sec:TWI}

%
%

Residual knowledge distillation aims to align the student backbone with the teacher backbone in a step-wise manner.
As shown in Fig.~\ref{fig:diagram}, we apply the pre-trained teacher classification head $\boldsymbol{h}_t$ to the MBRNet output $\{\hat{\boldsymbol{f}}_{k}\}_{k=1}^K$.
To prevent overfitting and enhance the task-specific learning capability, we incorporate a classification loss, $\mathcal{L}_{\text{cls}_k}$, for each branch.
The classification loss is defined as
\begin{equation}\label{eq:cls_k}
    \mathcal{L}_{\text{cls}_k} = \mathcal{L}_\text{CE}(\boldsymbol{y}, \sigma(\boldsymbol{h}_t(\hat{\boldsymbol{f}}_{k}))),
\end{equation}
where $\mathcal{L}_\text{CE}$ denotes the cross-entropy loss.
Building on this, the overall training loss $\mathcal{L}$ for ERA is formulated as
\begin{equation}
    \mathcal{L} = \mathcal{L}_{KD} + \sum_{k=0}^{K} s_k \cdot (\gamma\mathcal{L}_{\text{FD}_k} + \lambda\mathcal{L}_{\text{cls}_k}),
\end{equation}
where $\mathcal{L}_{KD}$ is the logit-based KD loss shown in Eq.~\eqref{eq:KD_logits}, $\gamma$ and $\lambda$ are weighting parameters balancing the importance of difference loss terms, and $s_k = \frac{1}{2^k}$ is a scaling factor that decreases as $k$ increases.
This scaling strategy prioritizes learning features closer to the student model's output in earlier branches, thereby improving training stability. 
Our experimental results validate the effectiveness of this approach, which will be further discussed in Sec.~\ref{sec:s_k}.

Leveraging a pre-trained teacher classification head $\boldsymbol{h}_t$ offers two significant advantages:
\textbf{1) Computational Efficiency.} Freezing the parameters of $\boldsymbol{h}_t$ eliminates the need for additional fine-tuning, thereby substantially reducing computational overhead. This is especially advantageous for tasks with computationally intensive heads, such as object detection.
\textbf{2) Knowledge Preservation.} As $\boldsymbol{h}_t$ is intrinsically aligned with the teacher model, it encapsulates rich knowledge of the teacher's latent space and learned representations~\cite{simKDCVPR2022}. Utilizing $\boldsymbol{h}_t$ to compute the logits of intermediate approximated features enables effective knowledge transfer from teacher to student. This process enhances the student model's ability to learn more meaningful and representative features.

\subsection{Inference Modes}\label{sec:Inf_Mode}

Since the auxiliary module MBRNet is introduced to assist the training of the student model, it enables three inference modes based on combinations of MBRNet and the student model: $S$-mode, $T$-mode, and $ST$-mode, Fig.~\ref{fig:diagram}(b).
Their key characteristics are summarized as follows.
\begin{itemize}

    \item \textbf{$\boldsymbol{S}$-mode:} 
    In this mode, inference operates in the same way as conventional methods. 
    The input image is processed solely by the student backbone and the student head $\boldsymbol{h}_s$ to produce the final prediction.
    The teacher head $\boldsymbol{h}_t$ and MBRNet are not used in this mode.

    \item \textbf{$\boldsymbol{T}$-mode:} 
    Unlike $S$-mode, $T$-mode excludes the student head $\boldsymbol{h}_s$. 
    The inference pipeline in this mode processes the input image through the student backbone, followed by MBRNet, and finally the teacher head $\boldsymbol{h}_t$ to generate the final prediction.
    The student head $\boldsymbol{h}_s$ is not utilized in this mode.

    \item \textbf{$\boldsymbol{ST}$-mode:} 
    This mode combines the strategies of both $S$ and $T$ modes.
    The input image is first processed by the student backbone to obtain $\boldsymbol{f}_s$, which is then fed into two separate paths.
    In the first path, $\boldsymbol{f}_s$ is passed through $\boldsymbol{h}_s$ to produce the student logits $\boldsymbol{p}_s$.
    In the second path, $\boldsymbol{f}_s$ is processed through MBRNet and $\boldsymbol{h}_t$ to generate the logits of mimicked teacher feature $\hat{\boldsymbol{p}}_t$.
    These two sets of logits are then merged to obtain the final prediction using the formula $\mu\boldsymbol{p}_s + (1-\mu)\hat{\boldsymbol{p}}_t$, where $\mu$ is a merging coefficient that controls the relative contributions of the student and teacher paths to the final output.
    
\end{itemize}

These three inference modes are designed to offer flexibility in balancing model performance and computational complexity by utilizing different combinations of ERA's components.
Specifically, the $S$-mode represents the baseline inference approach, featuring the simplest and most computationally efficient configuration. 
In contrast, the $T$-mode and $ST$-mode incorporate both the auxiliary MBRNet and the teacher head ($\boldsymbol{h_t}$), enhancing the model's capability at the cost of a slight increase in computational complexity.

This trade-off is validated by the experimental results presented in the subsequent section. 
In all experiments, the proposed ERA method is referred to as Ours-S', Ours-T', and `Ours-ST' when operating in the $S$-mode, $T$-mode, and $ST$-mode, respectively.

\begin{table*}[t]
    \as{1.2}
    \centering
    \caption{\textbf{Training strategies for image classification experiments.} \textit{BS}: batch size; \textit{LR}: learning rate; \textit{WD}: weight decay; \textit{LS}: label smoothing; \textit{EMA}: model exponential moving average; \textit{RA}: RandAugment~\cite{cubuk2020randaugment}; \textit{RE}: random erasing; \textit{CJ}: color jitter. For A1, B1, B2, and B3, the LR schedules are $\times 0.1$ at epochs $150, 180, 210$, $\times 0.1$ every $30$ epochs, $\times 0.1$ every $2.4$ epochs, and cosine, respectively.}
    
    \begin{tabular}{c|cccccccccc}
    \toprule
    Training Strategy & Dataset & Epochs & BS & LR & Optim & WD & LS & EMA  & Data Augmentation\\
    \midrule
    A1 & CIFAR-100 & 240 & 64 & 0.05 & SGD & $5\times10^{-4}$ & - & - & crop + flip \\	
    B1 & ImageNet & 100 & 256 & 0.1 & SGD & $1\times10^{-4}$ & - & - & crop, flip\\
    B2 & ImageNet & 450 & 768 & 0.048 & RMSProp & $1\times10^{-5}$ & 0.1 & 0.9999 & \{\textit{B1}\}, RA, RE\\
    B3 & ImageNet & 300 & 1024 & 5e-4 & AdamW & $5\times10^{-2}$ & 0.1 & - & \{\textit{B2}\}, CJ, Mixup, CutMix\\
    \bottomrule
    \end{tabular}
    \label{tab:cls_strategies}
\end{table*}

\begin{table*}[t]
    \as{1.2}
    \centering
    \caption{\textbf{Performance comparison on the CIFAR-100 dataset.} The Top-1 classification accuracy (\%) is used as the evaluation metric. The performance of the teacher and student models is indicated in parentheses following their model architecture.}

    \resizebox{0.9\linewidth}{!}{
    \begin{tabular}{l|ccc|ccc}
    \toprule
    Method & \multicolumn{3}{c|}{Same Architecture Design} & \multicolumn{3}{c}{Different Architecture Design}\\
		
    \midrule
    
    Teacher & WRN-40-2 (75.61) & ResNet-56 (72.34) & ResNet-32x4 (79.42) & ResNet-50 (79.34)   & ResNet-32x4 (79.42) & ResNet-32x4 (79.42) \\
    Student & WRN-40-1 (71.98) & ResNet-20 (69.06) & ResNet-8x4 (72.50)  & MobileNetV2 (64.60) & ShuffleNetV1 (70.50) & ShuffleNetV2 (71.82) \\
        
    \midrule
    KD~\cite{hinton2015distilling} & 73.54 & 70.66 & 73.33 & 67.35 & 74.07 & 74.45 \\
    FitNet~\cite{romero2014fitnets} & 72.24 & 69.21 & 73.50 & 63.16 & 73.59 & 73.54 \\
    DIST~\cite{huang2022knowledge} & 74.73 & 71.75 & 76.31 & 68.66 & 76.34 & 77.35 \\
    RKD~\cite{park2019relational}   & 72.22 & 69.61 & 71.90 & 64.43 & 72.28 & 73.21 \\
    CRD~\cite{tian2019contrastive} & 74.14 & 71.16 & 75.51 & 69.11 & 75.11 & 75.65 \\
    VID~\cite{ahn2019variational}   & 73.30 & 70.38 & 73.09 & 67.57 & 73.38 & 73.40 \\
    PKT~\cite{passalis2020probabilistic} & 73.45 & 70.34 & 73.64 & 66.52 & 74.10 & 74.69 \\
    AT~\cite{payingICLR2017}   & 72.77 & 70.55 & 73.44 & 58.58 & 71.73 & 72.73 \\
    \rowcolor{mygray}
    Ours-S & 74.63 & 71.85 & 76.47 & 69.01 & 76.50 & 77.44 \\
    \rowcolor{mygray}
    Ours-T & 74.71 & 71.90 & 76.77 & 69.11 & 76.58 & 77.54 \\
    \rowcolor{mygray}
    Ours-ST & \textbf{74.74} & \textbf{71.93} & \textbf{76.89} & \textbf{69.19} & \textbf{76.61} & \textbf{77.57}\\
    
    \bottomrule
    \end{tabular}
    }

    \label{tab:cifar100}
\end{table*}

\section{Experiment}\label{sec:experiments}

To evaluate the effectiveness and versatility of ERA, we conduct experiments on a wide range of tasks, including image classification, object detection, and semantic segmentation. We also carry out broad ablation studies to validate the effectiveness of ERA components.
%
%
%

\subsection{Image Classification}

\textbf{Settings.} 
Following the evaluation protocol in~\cite{huang2022knowledge}, teacher and student models with varying learning capabilities are used in the experiments. 
The training strategies for image classification on both the CIFAR-100~\cite{krizhevsky2009learning} and ImageNet~\cite{deng2009imagenet} datasets are summarized in Table~\ref{tab:cls_strategies}. 
On ImageNet, the strategies are categorized into two configurations: \textit{Baseline} and \textit{Advanced}.
In the \textit{Baseline} configuration, ResNet-18~\cite{he2016deep} and MobileNet V1~\cite{howard2017mobilenets} are chosen as student models, with ResNet-34 and ResNet-50 serving as their respective teacher models. This follows the training strategy B1, as described in~\cite{huang2022knowledge,zhao2022decoupled,chen2021distilling}.
For the \textit{Advanced} configuration, powerful teacher models, such as ResNet-50 and Swin-L~\cite{liu2021swin}, are employed alongside advanced training strategies (B2 and B3).
To minimize computational overhead, ERA is applied to the output features of backbone models after average pooling. This setup is compatible with both the $T$-mode and $ST$-mode inference settings.
For all image classification experiments, the hyper-parameters are consistently set as $\alpha = \lambda = \gamma = 1$ and $\beta = 2$. The Top-1 classification accuracy is reported as the evaluation metric.

\begin{table*}[ht]
    \as{1.2}
    \centering
    \caption{\textbf{Performance comparison on the ImageNet dataset.} ResNet-34 and ResNet-50~\cite{marcel2010torchvision} are used as the teacher networks, and the standard training strategy (\ie, B1) is adopted. The results of other methods quote the papers~\cite{chen2021distilling, huang2022knowledge,li2023curriculum,sun2024logit}. The performance of the teacher and student models is indicated in parentheses following their model architecture.}
    
    \begin{tabular}{cc|cccccccc}
        \toprule
        Student & Teacher & KD~\cite{hinton2015distilling} & Review~\cite{chen2021distilling} & DIST~\cite{huang2022knowledge} & CTKD~\cite{li2023curriculum} & LSKD~\cite{sun2024logit} & \cellcolor{mygray} Ours-S & \cellcolor{mygray}Ours-T & \cellcolor{mygray}Ours-ST \\
        \midrule
        ResNet-18 (69.76) & RetNet-34 (73.31) & 70.66 & 71.61 & 72.07 & 71.51 &  71.42 & \cellcolor{mygray}71.98 & \cellcolor{mygray}72.45 & \cellcolor{mygray}\textbf{72.49} \\
        MobileNet V1 (70.13) & ResNet-50 (76.16) & 70.68 & 72.56 & 73.24 & n/a & 72.18 & \cellcolor{mygray}73.25 & \cellcolor{mygray}73.89 & \cellcolor{mygray}\textbf{74.05} \\
        \bottomrule
    \end{tabular}

    \label{tab:baseline}
\end{table*}

\begin{table*}[ht]
    \as{1.2}
    \centering
    \caption{\textbf{Performance comparison on ImageNet with different teacher model architectures.} The learning strategy B1 is adopted. The performance of the teacher and student models is indicated in parentheses following their model architectures.}
    \label{tab:teachers}

    \begin{tabular}{ll|cccc}
    \toprule
    
    Student & Teacher & KD & \cellcolor{mygray} Ours-S & \cellcolor{mygray} Ours-T & \cellcolor{mygray} Ours-ST \\
    
    \shline
    
    \multirow{4}*{ResNet-18 (69.76)} & ResNet-34 (73.31) & 71.21 & \cellcolor{mygray}71.98 (\green{+0.77}) & \cellcolor{mygray}72.45 (\green{+1.24}) & \cellcolor{mygray}\textbf{72.49} (\green{+1.28}) \\
    
    ~ & ResNet-50 (76.13) & 71.35 & \cellcolor{mygray}71.92 (\green{+0.57}) & \cellcolor{mygray}72.31 (\green{+0.96}) & \cellcolor{mygray}\textbf{72.42} (\green{+1.07}) \\
    
    ~ & ResNet-101 (77.37) & 71.09 & \cellcolor{mygray}71.96 (\green{+0.87}) & \cellcolor{mygray}72.16 (\green{+1.07}) & \cellcolor{mygray}\textbf{72.45} (\green{+1.36}) \\
    
    ~ & ResNet-152 (78.31) & 71.12 & \cellcolor{mygray}71.99 (\green{+0.87}) & \cellcolor{mygray}72.36 (\green{+1.24}) & \cellcolor{mygray}\textbf{72.53} (\green{+1.41})\\
    
    \hline
    
    \multirow{3}*{ResNet-34 (73.31)} & ResNet-50 (76.13) & 74.73 & \cellcolor{mygray}74.87 (\green{+0.14}) & \cellcolor{mygray}75.07 (\green{+0.34}) & \cellcolor{mygray}\textbf{75.17} (\green{+0.44}) \\
    
    ~ & ResNet-101 (77.37) & 74.89 & \cellcolor{mygray}75.07 (\green{+0.18}) & \cellcolor{mygray}75.28 (\green{+0.39}) & \cellcolor{mygray}\textbf{75.48} (\green{+0.59}) \\
   
    ~ & ResNet-152 (78.31) & 74.87 & \cellcolor{mygray}75.13 (\green{+0.26}) & \cellcolor{mygray}75.34 (\green{+0.47}) & \cellcolor{mygray}\textbf{75.38} (\green{+0.51}) \\
   
   \bottomrule
   \end{tabular}
\end{table*}

\begin{table*}[ht]
    \as{1.2}
    \centering
    \caption{\textbf{Performance comparison on ImageNet of with strong teacher models and advanced learning strategies.} Experiments involving the Swin-T architecture adopt the B2 learning strategy, and others follow B3. $\dagger$: The model is trained following the protocol in~\cite{wightman2021resnet}. $\ddagger$: The model is pre-trained on ImageNet-22K.}
    
    \begin{tabular}{ll|cccccc}
    \toprule
    
    Student & Teacher & KD~\cite{hinton2015distilling} & RKD~\cite{park2019relational} & DIST~\cite{huang2022knowledge} & \cellcolor{mygray}Ours-S & \cellcolor{mygray}Ours-T & \cellcolor{mygray}Ours-ST \\

    \midrule
    
    ResNet-18 (73.4)  & \multirow{4}*{ResNet-50$^\dagger$ (80.1)}  & 72.6 & 72.9 & 74.5 & \cellcolor{mygray}74.7 & \cellcolor{mygray}75.1 & \cellcolor{mygray}\textbf{75.2} \\
    ResNet-34 (76.8) & ~ & 77.2 & 76.6 & 77.8 & \cellcolor{mygray}78.0 & \cellcolor{mygray}78.3 & \cellcolor{mygray}\textbf{78.5} \\
    MobileNetV2 (73.6) & ~ & 71.7 & 73.1 & 74.4 & \cellcolor{mygray}74.9 & \cellcolor{mygray}75.3 & \cellcolor{mygray}\textbf{75.4} \\
    EfficientNet-B0 (78.0) & ~ & 77.4 & 77.5 & 78.6 & \cellcolor{mygray}79.0 & \cellcolor{mygray}79.2 & \cellcolor{mygray}\textbf{79.4} \\

    \midrule
    
    ResNet-50 (78.5) & \multirow{2}*{Swin-L$^\ddagger$ (86.3)} & 80.0 & 78.9 & 80.2 & \cellcolor{mygray}80.5 & \cellcolor{mygray}80.7 & \cellcolor{mygray}\textbf{80.8} \\
    Swin-T (81.3) & ~ & 81.5 & 81.2 & 82.3 & \cellcolor{mygray}82.7 & \cellcolor{mygray}82.8 & \cellcolor{mygray}\textbf{83.0} \\
    
    \bottomrule
    \end{tabular}
    \label{tab:sota}
\end{table*}

\textbf{Results on CIFAR-100.}
%
From Table~\ref{tab:cifar100}, we can see that, when using the most lightweight and efficient $S$-mode inference, ERA (\ie. Ours-S) achieves results that are better or at least comparable to the benchmarks across all experimental settings. 
Furthermore, incorporating the MBRNet provides additional performance gains (as demonstrated by the results of Ours-T and Ours-ST), showcasing the flexibility and effectiveness of our TWI strategy during inference.

\textbf{Results on ImageNet.}
\textit{1) \textit{Baseline}:}
The experimental results on ImageNet under the \textit{Baseline} setting are presented in Table~\ref{tab:baseline}.
Ours-S is comparable to or slightly outperforms benchmark methods, highlighting ERA's effectiveness even in highly parameter-efficient scenarios.
In the $T$-mode, ERA (Ours-T) outperforms all competitors, achieving Top-1 accuracy of $72.45\%$ and $73.89\%$, which represent improvements of $1.79\%$ and $3.21\%$ over KD~\cite{hinton2015distilling}, respectively.
Additionally, the $ST$-mode further enhances performance, demonstrating ERA's ability to significantly reduce the performance gap between teacher and student networks.

In terms of computational efficiency, the auxiliary network MBRNet introduces minimal overhead. 
In the setting where the teacher and student models are respectively set to ResNet-34 and ResNet-18, MBRNet adds only 2.1M parameters, which accounts for 17.9\% of ResNet-18’s total 11.7M parameters.
Notably, applying ERA to the feature vectors after global average pooling increases the FLOPs by 0.067 and 0.068 GFLOPs for the Ours-T and Ours-ST modes, respectively. These represent modest increases of 3.69\% and 3.72\% relative to ResNet-18’s 1.814 GFLOPs.
This efficient design effectively minimizes the computational burden, making it particularly advantageous for deploying advanced deep learning models in resource-constrained environments.
Given that these ResNet variants share the same head and embedding dimensions, the increase in computational load remains consistent across setups.

For a more comprehensive evaluation, we further investigate the impact of different teacher model architectures.
As shown in Table~\ref{tab:teachers}, our method consistently outperforms KD~\cite{hinton2015distilling} across all inference modes, even when the teacher model is extended to more complex architectures, such as ResNet-50 and ResNet-101.
These results show that ERA performs robustly across varying teacher-student learning gaps, demonstrating its adaptability to network architectures.

\textit{2) \textit{Advanced}:}
To evaluate the adaptability of ERA under distillation gaps and learning strategies, we employ stronger teacher models and adopt the advanced learning strategies~\cite{huang2022knowledge}.
The results in Table~\ref{tab:sota} indicate that Ours-S consistently outperforms the state-of-the-art benchmark DIST in the \textit{Baseline} setting. Notably, this performance advantage becomes even more pronounced in $T$-mode and $ST$-mode.
These findings align with the results from Table~\ref{tab:teachers}, validating the compatibility of ERA with different training strategies.

\begin{table}[t]
    \as{1.2}
    \centering
    
    \caption{\textbf{Performance comparison of object detection on the validation set of COCO.} T.: Teacher; S.: Student; C.M.: Cascaded Mask.}
    
    \begin{tabular}{l|cccc}
    \toprule
    
    Method & AP & AP$_{S}$ & AP$_{M}$ & AP$_{L}$\\

    \midrule
    
    \multicolumn{5}{c}{\textit{Two-stage detectors}}\\
    T.: C.M. RCNN-X101 & 45.6 & 26.2 & 49.6 & 60.0\\
    S.: Faster RCNN-R50 & 38.4 & 21.5 & 42.1 & 50.3\\
    KD~\cite{hinton2015distilling} & 39.7 & 23.2 & 43.3 & 51.7\\
    FitNets~\cite{romero2014fitnets} & 39.9 & 23.4 & 43.4 & 52.0\\
    DIST~\cite{huang2022knowledge} & 40.4 & 23.9 & 44.6 & 52.6\\
    \rowcolor{mygray}
    Ours-S & 41.8 & \textbf{24.5} & 45.2 & 54.5 \\
    \rowcolor{mygray}
    Ours-T & 42.2 & 24.4 & 46.0 & 55.3 \\
    \rowcolor{mygray}
    Ours-ST & \textbf{42.4} & \textbf{24.5} & 46.1 & \textbf{55.7} \\

    \midrule

    \multicolumn{5}{c}{\textit{One-stage detectors}}\\
    T.: RetinaNet-X101 & 41.0 & 23.9 & 45.2 & 54.0\\
    S.: RetinaNet-R50 & 37.4 & 20.0 & 40.7 & 49.7\\
    KD~\cite{hinton2015distilling} & 37.2 & 20.4 & 40.4 & 49.5\\
    FitNets~\cite{romero2014fitnets} & 37.5 & 20.9 & 40.9 & 51.1 \\
    DIST~\cite{huang2022knowledge} & 39.8 & 22.0 & 43.7 & 53.0\\
    \rowcolor{mygray}
    \customstrut{1ex}Ours-S & 40.6 & 23.1 & 44.6 & 53.9 \\
    \rowcolor{mygray}
    \customstrut{1ex}Ours-T & 40.7 & 23.1 & 45.0 & 54.5 \\
    \rowcolor{mygray}
    \customstrut{1ex}Ours-ST & \textbf{40.8} & \textbf{23.3} & \textbf{45.2} & \textbf{54.7} \\
    
    \bottomrule
    
    \end{tabular}
    \label{tab:det}
\end{table}

\subsection{Object Detection}

\textbf{Settings.}
Following DIST~\cite{huang2022knowledge}, we evaluate ERA on the COCO dataset~\cite{lin2014microsoft} using both two-stage and one-stage object detection frameworks, with Cascade Mask R-CNN~\cite{cai2019cascade} and RetinaNet~\cite{lin2017focal} serving as the detection heads, respectively.
For the teacher and student model, we use ResNeXt-101~\cite{xie2017aggregated} (X101) and ResNet-50~\cite{he2016deep} (R50), respectively.
For the training strategies, we follow the protocols proposed in prior studies~\cite{yang2022focal, huang2022knowledge}. 
We apply ERA to the neck module of the detection framework, implemented using the FPN~\cite{lin2017feature}. 
Specifically, the proposed MBRNet is integrated into each FPN branch to enhance the distillation process across multiple feature scales. 
In the $ST$-mode, proposals generated by the student and teacher RPNs are merged and passed through the teacher detection head to produce the final results.

\textbf{Results.}
The experimental results in Table~\ref{tab:det} demonstrate the performance improvements achieved by our method.
By retaining the inference architecture of FitNets~\cite{romero2014fitnets}, Ours-S improves the Average Precision (AP) by $1.7$ on two-stage frameworks and $3.0$ on one-stage frameworks.
With the integration of MBRNet, the detection accuracy is further enhanced, achieving $42.4$ AP on the two-stage framework and $40.8$ AP on the one-stage framework, as shown by the results of Ours-ST.
These results highlight the effectiveness and adaptability of ERA in distilling object detection models.

Notably, our method in $S$-mode (\ie, Ours-S) matches the computational cost of DIST~\cite{huang2022knowledge} while enjoying higher performance. Specifically, Ours-S achieves $1.4$ AP and $0.8$ AP gains over DIST on two-stage and one-stage detectors, respectively. This improvement stems from our integration of MBRNet with ERA training, in contrast to DIST.

\begin{table}[t]
    \as{1.2}
    \centering
    \caption{\textbf{Performance comparison of semantic segmentation on the validation set of Cityscapes dataset.} T.: Teacher; S.: Student.}
    \begin{tabular}{l|c}
    
        \toprule
        Method & mIoU (\%)\\
        \midrule
        T.: DeepLabV3-R101 & 78.07\\
        
        \midrule
        
        S.: DeepLabV3-R18 & 74.21\\
        DIST~\cite{huang2022knowledge} & 77.10\\
        SKD~\cite{liu2019structured} & 75.42\\
        IFVD~\cite{wang2020intra} & 75.59\\
        CWD~\cite{shu2021channel} & 75.55\\
        CIRKD~\cite{yang2022cross} & 76.38\\
        Auto-KD~\cite{li2023automated} & 77.35 \\
        \rowcolor{mygray}Ours-S & 77.60 \\
        \rowcolor{mygray}Ours-T & \textbf{77.80} \\
        \rowcolor{mygray}Ours-ST & 77.75 \\

        \midrule
        
        S.: PSPNet-R18 & 72.55\\
        DIST~\cite{huang2022knowledge} & 76.31 \\
        SKD~\cite{liu2019structured} & 73.29\\
        IFVD~\cite{wang2020intra} & 73.71\\
        CWD~\cite{shu2021channel} & 74.36\\
        CIRKD~\cite{yang2022cross} & 74.73\\
        Auto-KD~\cite{li2023automated} & 76.25 \\
        \rowcolor{mygray}Ours-S & 76.57 \\
        \rowcolor{mygray}Ours-T & \textbf{76.83} \\
        \rowcolor{mygray}Ours-ST & 76.75 \\
    \bottomrule
    
    \end{tabular}
    \label{tab:seg}
\end{table}

\subsection{Semantic Segmentation}

\textbf{Settings.} 
Following the protocol in~\cite{yang2022cross}, we adopt DeepLabV3~\cite{chen2018encoder} with ResNet-101~\cite{he2016deep} (R101) as the teacher network and conduct experiments on the Cityscapes dataset~\cite{cityscapesCVPR2016}..
For the student networks, we utilize frameworks, including DeepLabV3 and PSPNet~\cite{zhao2017pyramid}, both based on the ResNet-18~\cite{he2016deep} (R18) backbone model, to evaluate the effectiveness of our method and its generalizability across architectures.
Note that the ERA-based feature alignment is performed on the features extracted after the atrous spatial pyramid pooling (ASPP) module in DeepLabV3.

\begin{table*}[t]
\as{1.2}
\centering
\caption{\textbf{Ablation study on the proposed ERA approach and TWI training strategy.} The results are obtained from the ImageNet classification task using the B1 protocol, with ERA operating in S-mode. For all knowledge distillation methods, the student and teacher models are ResNet-18 and ResNet-34, respectively.}
\begin{tabular}{lcccccc}
\toprule
Methods & $\mathcal{L}_{KL}$ & $\boldsymbol{h}_s$ & $\boldsymbol{h}_t$ & Feature Alignment & $\mathcal{L}_{\text{cls}_k}$ & Acc. (\%) \\
\midrule
KD         & \checkmark & \checkmark & & & & 71.21 \\
FitNet     & \checkmark & \checkmark & & $\mathcal{L}_{FD}$ & & 71.29 \\
TAKD       & \checkmark & \checkmark & & Assistant Net (w/o $\mathcal{L}_{FD}$) & & 71.50 \\
TAKD*      & \checkmark & \checkmark & & Assistant Net (w/ $\mathcal{L}_{FD}$) & & 71.53 \\
ResErrKD      & \checkmark & \checkmark & \checkmark & Assistant Net (w/o $\mathcal{L}_{FD}$) & & 71.46 \\
SimKD      & \checkmark &  \checkmark & \checkmark & $\mathcal{L}_{FD}$ & & 71.32 \\
\rowcolor{mygray}
TWI        & \checkmark & \checkmark & \checkmark & $\mathcal{L}_{FD}$ & & 71.34 \\
\rowcolor{mygray}
ERA$_{\psi}$ & \checkmark & \checkmark & \checkmark & MBRNet (w/o $\mathcal{L}_{FD}$) & & 71.56 \\
\rowcolor{mygray}
ERA$_{\zeta}$ & & \checkmark & \checkmark & MBRNet (w/ $\mathcal{L}_{FD}$) & & 71.54 \\
\rowcolor{mygray}
ERA$_{\iota}$ &  & \checkmark & \checkmark & MBRNet (w/o $\mathcal{L}_{FD}$) & \checkmark & 71.16 \\
\rowcolor{mygray}
ERA$_{\S}$  & \checkmark & \checkmark & \checkmark & MBRNet (w/ $\mathcal{L}_{FD}$) & & 71.60 \\
\rowcolor{mygray}
ERA$_{\dagger}$ & \checkmark & \checkmark & \checkmark & MBRNet (w/ $\mathcal{L}_{FD}$) & & 71.75 \\
\rowcolor{mygray}
ERA$_{\tau}$ &  & \checkmark & \checkmark & MBRNet (w/ $\mathcal{L}_{FD}$) & \checkmark & 71.93 \\
\rowcolor{mygray}
ERA         & \checkmark & \checkmark & \checkmark & MBRNet (w/ $\mathcal{L}_{FD}$) & \checkmark & \textbf{71.98} \\
\bottomrule
\end{tabular}
\label{tab:MBRNet_TWI_Ablation}
\end{table*}

\textbf{Results.} 
As shown in Table~\ref{tab:seg}, even with the most lightweight $S$-mode configuration, ERA (\ie, Ours-S) achieves competitive mIoU scores of 77.60 and 76.57 on DeepLabV3-18 and PSPNet-R18, respectively. 
These results outperforms the previous state-of-the-art method, Auto-KD~\cite{li2023automated}, by 0.35 and 0.32 mIoU.
%
The performance advantage of our method is further amplified when using the $T$-mode configuration. In this mode, the mIoU scores increase by 0.20 and 0.26 on DeepLabV3-18 and PSPNet-R18, respectively, narrowing the gap to the teacher model to 0.27 and 1.24 mIoU.

\subsection{Comparison Analysis}

\textbf{Comparing with KD Approaches with Assistant Networks}
We compare ERA with ResErrKD~\cite{gao2021residual} and TAKD~\cite{mirzadeh2020improved}, which similarly utilize auxiliary networks to reduce the learning gap between teacher and student models.
In summary, the advantages of ERA over ResErrKD and TAKD are three-fold:

\textit{1) Efficient Multi-step Approximation.}
Unlike ResErrKD, which is limited to a two-step approximation, our MBRNet provides a flexible $K$-step solution.
Specifically, ResErrKD uses a single assistant network to transfer the teacher's knowledge to the student in two steps: from $\mathcal{P}_0\boldsymbol{f_s}$ to $\boldsymbol{\hat{f}}_1$ and then to $\boldsymbol{\hat{f}}_t$.
In contrast, MBRNet introduces $K$ intermediary branches, enabling a knowledge approximation over $K\!+\!1$ steps: from $\mathcal{P}_0\boldsymbol{f_s}$ to $\boldsymbol{\hat{f}}_1, \cdots, \boldsymbol{\hat{f}}_K$, and finally to $\boldsymbol{\hat{f}}_t$.
This stepwise approach is validated to be a more effective and scalable solution for bridging the gap between teacher and student models.
As for TAKD, which requires multiple iterative distillation steps, MBRNet completes the process in a single step. 
With its multiple branches, our MBRNet effectively replaces TAKD's intermediate TAs, enabling simultaneous training of both MBRNet and the student network.

\textit{2) Simple Implementation.} 
To avoid additional computational overhead, ResErrKD splits the original student model into two sub-networks: one as the new student, the other as its assistant. 
TAKD, on the other hand, requires constructing a sequence of teacher assistants with progressively lower capacities than the teacher, gradually approaching the student with multiple training stages.
%
In contrast, MBRNet adopts a simpler structure, achieving better performance (as shown by the ERA$_{\S}$ results in Table~\ref{tab:MBRNet_TWI_Ablation}) with minimal implementation effort.

\textit{3) Enhanced Effectiveness.}
ERA significantly outperforms ResErrKD and TAKD in both performance and computational efficiency. 
As shown in Table~\ref{tab:MBRNet_TWI_Ablation}, by replacing the assistant network with MBRNet, ERA$_{\S}$ achieves a $0.14$ improvement over ResErrKD and a $0.10$ improvement over TAKD in classification accuracy, all without incurring extra computational overhead. 
When combined with pre-trained $\boldsymbol{h}_t$, ERA achieves even more substantial gains, surpassing ResErrKD by $0.52$ and TAKD by $0.48$. 
These improvements underscore the advantages of MBRNet in delivering superior KD performance with minimal complexity.

\begin{table}[t]
    \as{1.2}
    \centering
    \caption{Performance comparison of ERA with stronger baseline methods DKD~\cite{zhao2022decoupled} and ReviewKD~\cite{chen2021distilling}. The teacher model is ResNet-34, while the student model is ResNet-18.}
    
    \begin{tabular}{cccc}
    \toprule
        \multirow{2.3}{*}{ReviewKD} & \multicolumn{3}{c}{ReviewKD+ERA} \\
        \cmidrule{2-4}
        {} & Ours-S & Ours-T & Ours-ST \\
        71.61 & \cellcolor{mygray}72.05 & \cellcolor{mygray}\textbf{72.53} & \cellcolor{mygray}72.50 \\
        \midrule
        \multirow{2.3}{*}{DKD} & \multicolumn{3}{c}{DKD+ERA} \\
        \cmidrule{2-4}
        {} & Ours-S & Ours-T & Ours-ST \\
        71.70 & \cellcolor{mygray}72.11 & \cellcolor{mygray}72.60 & \cellcolor{mygray}\textbf{72.63} \\
        \bottomrule
    \end{tabular}
    \label{tab:dkd}
\end{table}

\subsection{Ablation Studies}\label{sec:ablations}

\textbf{Teacher Weight Integration (TWI).} 
Building on FitNet, SimKD~\cite{simKDCVPR2022} further re-uses the classifier from the pre-trained teacher model (\ie, adopting the student backbone with a frozen teacher head) for student model distillation, but this modification only provides a minor additional gain of $0.03$.
TWI slightly differs with SimKD, it can be regarded as conducting FitNet~\cite{romero2014fitnets} while adopting teacher head $\boldsymbol{f}_t$ when evaluation.
With the negligible performance improvement against SimKD, we can conclude that TWI cannot solely improves the performance.
Compared with TWI and ERA, we know that when adopting parameterized MBRNet as the feature alignment module and together with utilizing $\mathcal{L}_{cls_k}$, it can boost the performance significantly.

\textbf{MBRNet.}
To evaluate MBRNet's effectiveness in approximating residual knowledge, we thoroughly explore the impact of each component.
As indicated by Table~\ref{tab:MBRNet_TWI_Ablation}, the results show that ERA$_{\psi}$ achieves performance comparable to ERA$_{\zeta}$. However, $\mathcal{L}_{cls_k}$ cannot be used alone without $\mathcal{L}_{KL}$ or $\mathcal{L}_{FD}$, because it is designed to prevent overfitting in MBRNet.
Additionally, ERA$_{\tau}$, trained with both $\mathcal{L}_{FD}$ and $\mathcal{L}_{cls_k}$, shows significant improvement.
Besides, ERA${\S}$ surpasses FitNet by a margin of $0.31$ in classification accuracy, demonstrating that MBRNet can serve as a plug-and-play replacement for MSE, even without leveraging $\boldsymbol{h}_t$, and consequently without applying $\mathcal{L}_{\text{cls}_k}$ (as defined in Eq.~\eqref{eq:cls_k}).
Furthermore, progressively incorporating $\boldsymbol{h}_t$ and $\mathcal{L}_{\text{cls}_k}$ results in the models ERA${\dagger}$ and ERA, which achieve additional performance gains of $0.15$ and $0.38$, respectively. 
These improvements are significantly larger than those yielded by FitNet and SimKD ($0.08$ and $0.03$), underscoring the effectiveness of ERA based on MBRNet.

\begin{figure}[t]
    \centering
    \includegraphics[width=0.9\linewidth]{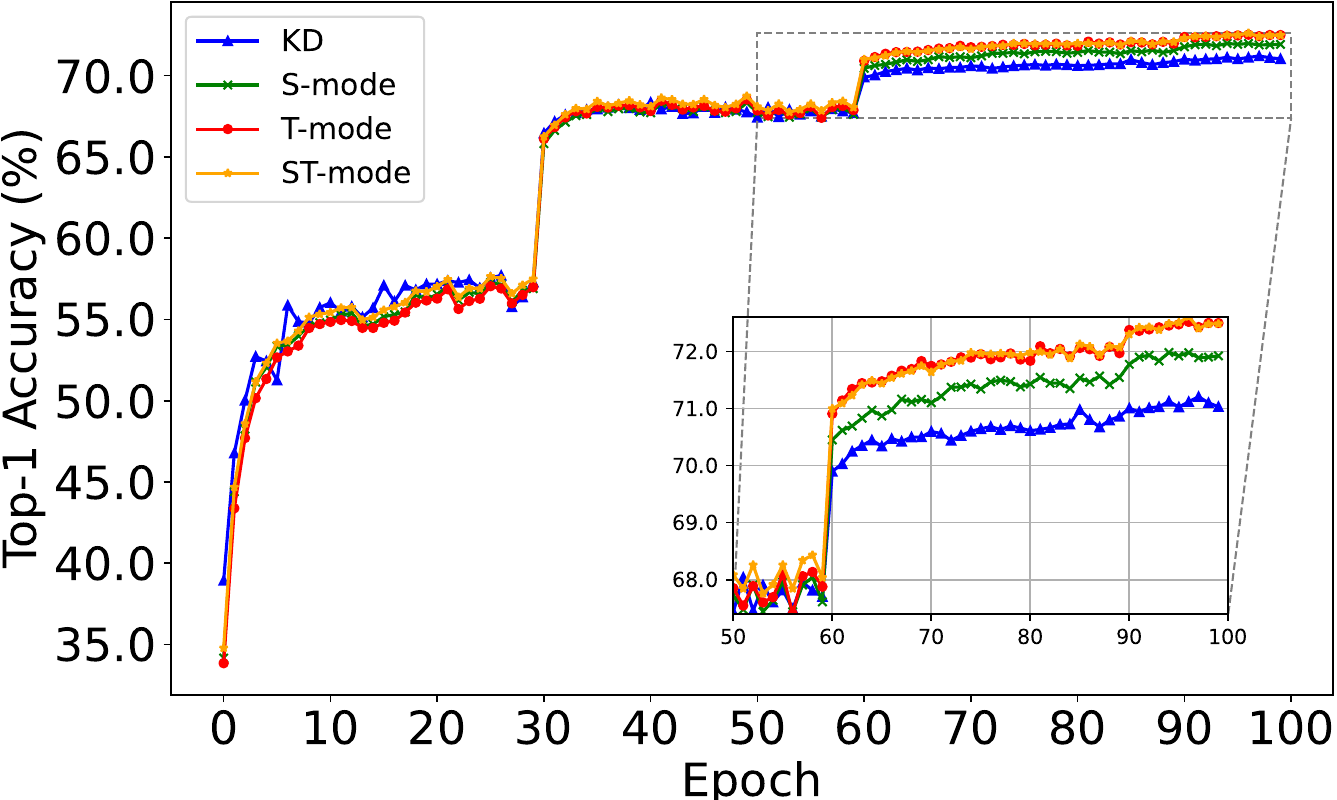}
    \caption{\textbf{Performance of ERA of inference modes and against KD~\cite{hinton2015distilling}.} Results are based on the ImageNet classification task, with ResNet-34 (R34) as the teacher model and ResNet-18 (R18) as the student model.}
    \label{fig:acc_modes_KD}
\end{figure}

\textbf{Branch Number and Inference Mode.} 
%
We analyze the impact of inference modes and the number of branches included in the MBRNet. 
Fig.~\ref{fig:acc_modes_KD} presents the training dynamics of ERA in $S$-mode, $T$-mode, and $ST$-mode throughout the entire training process for image classification, and compares them with the benchmark method KD~\cite{hinton2015distilling}. 
Starting from epoch $60$, when the final learning rate decay begins under the B1 strategy, ERA consistently outperforms KD. 
This result highlights the effectiveness of ERA. Furthermore, $T$-mode and $ST$-mode achieve higher accuracy than $S$-mode, demonstrating the benefits of our TWI strategy.

\begin{figure}[t]
    \centering
    \includegraphics[width=0.9\linewidth]{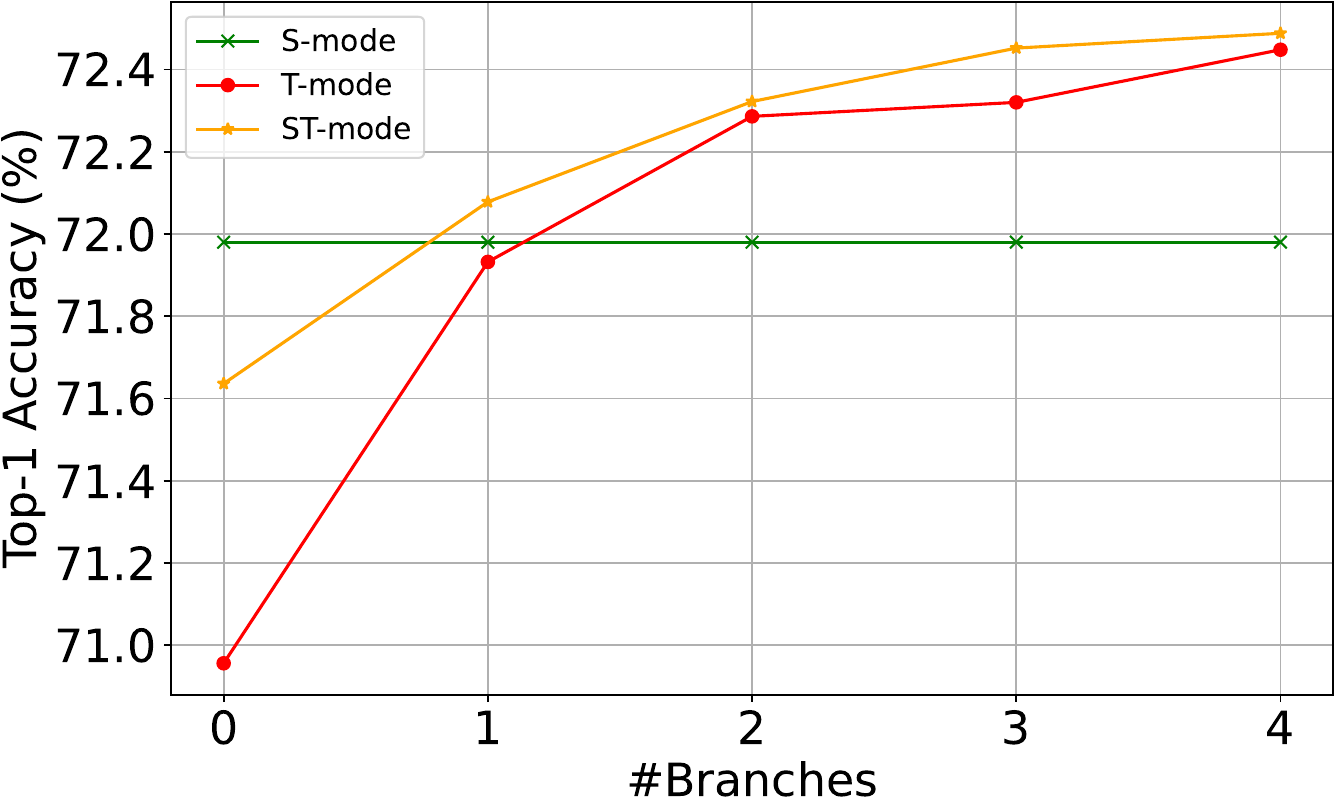}
    \caption{\textbf{Performance of ERA under numbers of MBRNet branches.} It is evaluated on a distilled ResNet-18 (teacher ResNet-34, $K=4$ by default), progressively incorporating branches. As an example, \#Branches=2 denotes that only the 0-th and 1-st branches are employed.}
    \label{fig:number_branches}
\end{figure}

We further analyze how the number of branches in MBRNet affects the performance of ERA. 
As shown in Fig.~\ref{fig:number_branches}, $T$-mode and $ST$-mode exhibit consistent performance improvements as the number of inference branches increases from 0 to 4. 
This demonstrates that MBRNet enables a progressive approximation of the `residual knowledge' between teacher and student models. 
Moreover, as the number of branches increases, each approximation step becomes smaller and more manageable, progressively reducing the knowledge gap and leading to better overall performance.

\begin{table}[t]
    \as{1.2}
    \centering
    
    \caption{\textbf{Performance of the teacher network with learnable/frozen classifier weights.} Stu.: Student; Tea.: Teacher.}
    
    \begin{tabular}{l|c|ccccc}
    
    \toprule
    
    Stu. (Tea.) & KD~\cite{hinton2015distilling} & $\boldsymbol{h}_t$ & \cellcolor{mygray} Ours-S & \cellcolor{mygray}Ours-T & \cellcolor{mygray}Ours-ST \\
    
    \midrule
    
    \multirow{2}*{R18 (R34)}  & \multirow{2}*{71.21} & Learnable & \cellcolor{mygray}71.72 & \cellcolor{mygray}72.37 & \cellcolor{mygray}\textbf{72.40}\\
    {} & {}  & Frozen & \cellcolor{mygray}71.98 & \cellcolor{mygray}72.45 & \cellcolor{mygray}\textbf{72.49} \\
    \hline
    \multirow{2}*{MBV1 (R50)}  & \multirow{2}*{76.16} & Learnable & \cellcolor{mygray}73.01 & \cellcolor{mygray}73.75 & \cellcolor{mygray}\textbf{73.96}\\
    {} & {} & Frozen & \cellcolor{mygray}73.25 & \cellcolor{mygray}73.89 & \cellcolor{mygray}\textbf{74.05} \\
    
    \bottomrule
    
    \end{tabular}
    \label{tab:learnable_ht}
\end{table}

\textbf{Learnable Teacher Head Weights.}
In ERA, the teacher's pre-trained classifier ($\boldsymbol{h}_t$) is reused to predict logits for each MBRNet branch, with its weights kept frozen throughout training to preserve the knowledge acquired during pre-training. 
In this subsection, we explore whether making $\boldsymbol{h}_t$ learnable can improve the performance of the distilled model. 
Experimental results for various teacher-student architecture pairings are presented in Table~\ref{tab:learnable_ht}.

The results clearly show that the frozen teacher classifier consistently outperforms its learnable counterpart. 
This advantage may stem from the frozen weights retaining the rich knowledge embedded in $\boldsymbol{h}_t$ while avoiding potential noise or degradation introduced during training. 
Specifically, in $S$-mode inference, the performance gap between frozen and learnable teacher heads is $0.26$ for the R18 (R34) distillation setup. 
This gap narrows to $0.08$ in $T$-mode and $0.09$ in $ST$-mode, with similar trends observed in the MBV1 (R50) distillation setup. 
These results indicate that while a learnable teacher head slightly reduces the performance gap in $T$-mode and $ST$-mode, it remains marginally inferior to frozen weights overall.

%
%
%
%

\begin{table}[t]
    \as{1.2}
    \centering
    
    \caption{\textbf{Performance comparison of different scheduling strategies of $s_k$.} `N/A' refers to the `NaN (Not A Number) Error'.}
    
    \begin{tabular}{l|ccc}
    \toprule
    Scheduling Strategies & Ours-S & Ours-T & Ours-ST \\
    \midrule
    $s_k=1$       & 71.84 & 72.17  & 72.23 \\
    \midrule
    $s_k=k$       & N/A   & N/A    & N/A \\
    $s_k=2^k$     & N/A   & N/A    & N/A \\
    \midrule
    $s_0=1$, $s_{k\geq1}=1e^{-6}$ & 71.31 & 71.36  & 71.37 \\
    $s_k=1$, $s_{k\geq1}$ linear decay & 71.40 & 71.85 & 71.91 \\
    \midrule
    $s_k=1/(1+k)$ & 71.88 & 72.36  & 72.42 \\
    $s_k=1/2^{k}$ & \textbf{71.98} & \textbf{72.45} & \textbf{72.49} \\
    \bottomrule
    \end{tabular}
    \label{tab:wscale}
\end{table}

\textbf{Layer-wise Decaying Weighting Scheme}\label{sec:s_k}
In Sec.~\ref{sec:TWI}, we introduced the overall training objective for ERA, where the adaptive weighting parameters $\{s_k\}_{k=0}^K$ are designed to provide layer-dependent balancing for the objective functions. 
This design reduces the impact of the MBRNet branch with larger indices $k$, \ie, farther from the student model's output $\Delta\hat{\boldsymbol{f}}_0$, allowing higher error tolerance in feature alignment and class prediction. 
As a result, it is theoretically expected to improve both the convergence and stability of the training process.
In this subsection, we assess the effectiveness of this Layer-wise Decaying Weighting Scheme and compare it with three categories of alternative strategies: (i) constant weighting, (ii) increasing weighting, and (iii) heavily biased setting where the first-branch weight $s_0$ far exceeds those of all remaining branches.

As shown in Table~\ref{tab:wscale}, weighting schemes with increasing values of $s_k$ relative to $k$ (\ie, $s_k\!=\!k$ and $s_k\!=\!2^k$) lead to unstable training, resulting in NaN (\ie, Not a Number) loss values across all inference settings. 
While using a constant $s_k$ resolves the instability, it performs worse than schemes with decreasing weights. 
For the biased scheduling approach, we set $s_0=1$ while assigning all other weights (\textit{i.e.}, $s_{k\geq1}$) a constant value of $1e^{-6}$, which is effectively close to zero.
This scheduling strategy results in a significant performance drop, reaching the level of the vanilla FitNet baseline.
We further tried a dynamic schedule in which we initialize all weights with $s_k=1$ and linearly decay the auxiliary ones ($k\ge1$) to 0 during training.
Table~\ref{tab:wscale} shows that the linear-decay schedule surpasses the aforementioned setting because the auxiliary branches are active at the beginning of training; however, once their weights fade to zero it falls short of the constant-weight baseline ($s_k = 1$).

For the decreasing strategy, we examine two implementations: $s_k\!=\!1/(1\!+\!k)$ (linear decay) and $s_k\!=\!1/2^k$ (exponential decay). 
Experimental results demonstrate that the exponential decay scheme consistently outperforms the linear decay approach across all inference modes. 
Consequently, we adopt the exponential decay scheme as the default strategy in our experiments while acknowledging the potential for further optimization through customized adjustments.

\textbf{ERA with Stronger Baselines}
To further investigate the potential of ERA training, we apply it to stronger baselines, namely ReviewKD~\cite{chen2021distilling} and DKD~\cite{zhao2022decoupled}. Specifically, for ReviewKD, MBRNet is integrated after the Pyramid Pooling operation in each HCL block. As shown in , even in the $S$-mode, which incurs no additional computational overhead, ERA training enhances performance by $0.44$ for ReviewKD and $0.31$ for DKD. The $T$- and $ST$-mode deliver further performance improvements, highlighting the effectiveness of ERA across inference modes.

\begin{table}[t]
    \as{1.2}
    \centering
    
    \caption{\textbf{Performance Comparison of MBRNet Variants.} $K$ represents the number of approximating steps and $m$ denotes the number of `FC $\rightarrow$ BN $\rightarrow$ ReLU' blocks. The teacher model is ResNet-34, while the student model is ResNet-18.}
	
    \begin{tabular}{cc|ccc}
    \toprule
    $m$ & $K$ & Ours-S & Ours-T & Ours-ST \\
    \midrule
    1 & 1 & 71.31  & 71.40  & 71.43 \\
    1 & 2 & 71.50  & 71.69  & 71.80 \\
    1 & 3 & 71.55  & 71.74  & 71.84 \\
    \midrule
    2 & 2 & 71.64  & 71.87  & 72.01 \\
    2 & 3 & 71.80  & 72.34  & 72.39 \\
    2 & 4 & \textbf{71.98}  & 72.45  & 72.49 \\
    2 & 5 & 71.91  & \textbf{72.47}  & \textbf{72.50} \\
    \midrule
    3 & 4 & 72.89  & 72.40  & 72.45 \\
    \bottomrule
    \end{tabular}
    
    \label{tab:mbrnet}
\end{table}

\subsection{Hyper-parameter Selection}
%

\textbf{MBRNet Network Parameters.}
The architecture of MBRNet is defined by two hyper-parameters: the number of approximation steps ($K$) and the number of `FC $\rightarrow$ BN $\rightarrow$ ReLU' blocks ($m$) in each branch.
We analyze the impact of these hyper-parameters on the performance of the proposed ERA method, with the results summarized in Table~\ref{tab:mbrnet}.

It can be seen that increasing $K$ from 1 to 4 results in significant performance improvements across all inference modes. 
However, further increasing $K$ to 5 yields only marginal gains in the $T$- and $ST$-inference modes (improvements of $0.02$ and $0.01$, respectively) and a slight accuracy drop in the $S$-mode (from $71.98$ to $71.91$). 
This decline can be attributed to the accumulation of approximation errors introduced by additional steps.
Similarly, increasing $m$ from 1 to 2 improves performance across all inference modes, primarily due to the enhanced model capacity. 
While increasing $m$ to 3 further boosts $S$-mode performance (from $71.98$ to $72.89$), it results in performance degradation in the $T$- and $ST$-modes.
Based on these findings, we set $K = 4$ and $m = 2$ to achieve a balanced trade-off between model complexity and performance.

\begin{table}[t]
    \as{1.2}
    \centering
    \caption{\textbf{Performance comparison of different combinations of $\gamma$ and $\lambda$.} By default, we set $\alpha=1$ and $\beta=2$ following~\cite{huang2022knowledge}.}
	
    \begin{tabular}{cc|ccc}
        \toprule
        $\gamma$ & $\lambda$ & Ours-S  & Ours-T  & Ours-ST  \\
        \midrule
        0.1 & 0.1 & 71.60  & 72.11  & 72.29 \\
        0.5 & 0.1 & 71.65  & 72.17  & 72.35 \\
        1.0 & 0.1 & 71.73  & 72.21  & 72.33 \\
        2.0 & 0.1 & 71.62  & 72.14  & 72.30 \\
        \midrule
        1.0 & 0.5 & 71.88  & 72.44  & 72.60 \\
        1.0 & 1.0 & \textbf{71.98}  & \textbf{72.45}  & \textbf{72.49} \\
        1.0 & 2.0 & 71.83  & 72.33  & 72.43 \\
        \bottomrule
    \end{tabular}
\label{tab:gamma_lambda}
\end{table}

\textbf{Loss Balancing Parameters.}
To reduce the number of hyperparameters requiring tuning, we follow the protocol outlined in DIST~\cite{huang2022knowledge}. 
Specifically, we fix $\alpha=1$ and $\beta=2$ to construct the KD baseline, focusing only on tuning $\gamma$ and $\lambda$ to evaluate their influence.
As shown in Table~\ref{tab:gamma_lambda}, our method achieves optimal performance across all inference modes when $\gamma=1$ and $\lambda=1$, which we adopt for all experiments. 
This result underscores the equal importance of feature alignment and task-specific objectives (\eg, classification).

\section{Limitation}\label{sec:limitation}
While the proposed ERA method demonstrates significant advantages across various tasks, there are certain limitations that deserve further exploration. 
First, although the algorithm design is conceptually straightforward, its practical implementation introduces complexity due to the multi-branch structure of MBRNet and the interdependence between target features and predictions from prior stages. This complexity may pose challenges in deployment and scalability.
Second, the choice of the hyper-parameter $K$ (\textit{i.e.}, the number of network branches in MBRNet) is task-specific and affects the trade-off between representation accuracy and computational cost. While we set $K=4$ across all experiments and achieved satisfactory results, further fine-tuning $K$ for specific tasks may yield better performance but also increases computational demands.
Finally, due to its reliance on feature distillation, the ERA method encounters challenges when applied to the knowledge distillation of Large Language Models (LLMs), where accessing intermediate activations is computationally and memory-intensive. In such cases, logits-based distillation remains a more practical alternative.
Despite these limitations, the proposed method serves as a promising foundation for further research, and future work could focus on addressing these challenges to enhance its applicability and efficiency.

\section{Conclusion}\label{sec:conclusion}

This study introduces the Expandable Residual Approximation (ERA) method for knowledge distillation, aimed at addressing the substantial disparity in learning capabilities between teacher and student models.
To achieve this, ERA employs a multi-branched residual network (MBRNet), which captures the residual knowledge between the teacher and student models. MBRNet facilitates the step-by-step optimization of feature approximation by systematically refining the student's output.
Additionally, ERA incorporates a Teacher Weight Integration (TWI) strategy, which further mitigates the capability gap by reusing the teacher model's classifier weights to enhance the approximation results produced by different branches of MBRNet.
Comprehensive experiments demonstrate that ERA consistently achieves state-of-the-art performance across datasets and tasks, validating the robustness and generalizability of the proposed approach.

\bibliographystyle{IEEEtran}
\bibliography{main}

\begin{IEEEbiography}[{\includegraphics[width=1in,height=1.25in,clip,keepaspectratio]{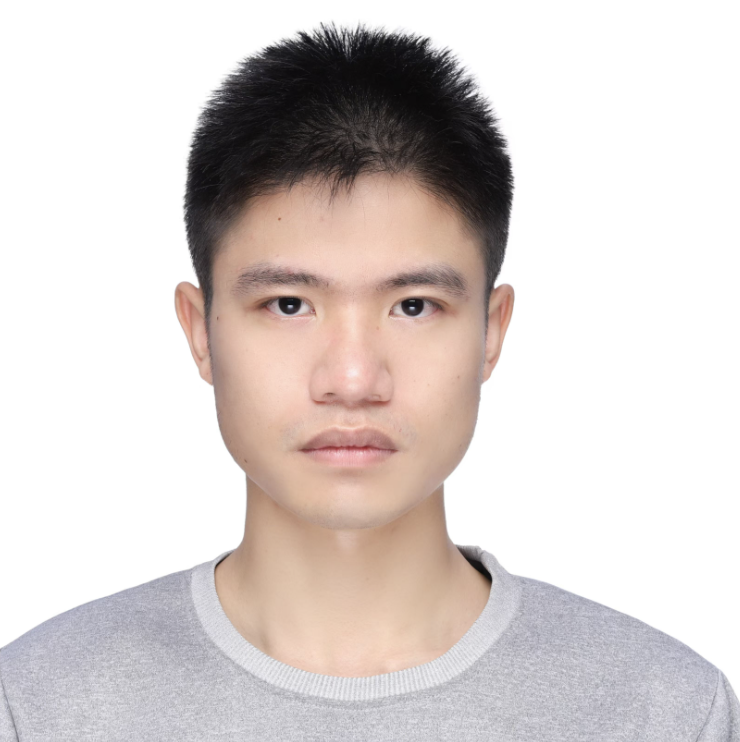}}]{Zhaoyi Yan}  received the Ph.D. degree in Computer Science from Harbin Institute of Technology, China, in 2021. His research interests include deep learning, image inpainting, crowd counting, knowledge distillation and Large Language Models. He has published more than 10 papers in conferences and journal including CVPR, ICCV, ECCV, AAAI, TNNLS and TCSVT.
\end{IEEEbiography}

\begin{IEEEbiography}[{\includegraphics[width=1in,height=1.25in,clip,keepaspectratio]{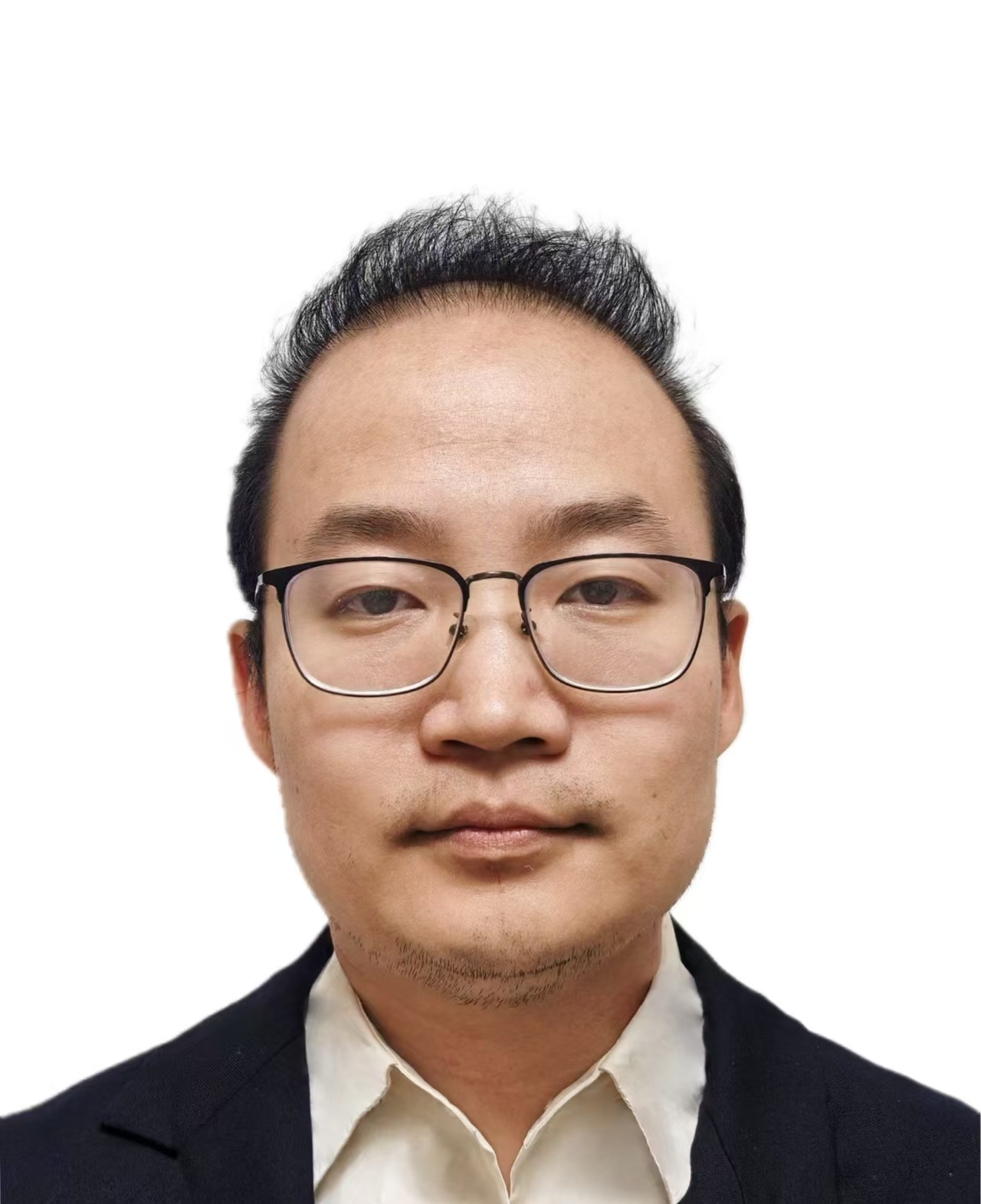}}] {Binghui Chen} received the B.E. and Ph.D. degrees in telecommunication engineering from the Beijing University of Posts and Telecommunications (BUPT), Beijing, China, in 2015 and 2020, respectively. His research interests include computer vision, deep learning, face recognition, deep embedding learning, crowd-counting, object detection and machine learning. He has published more than 20 papers in conferences and journal including CVPR, ICCV, NeurIPS, AAAI and TNNLS.
\end{IEEEbiography}

\begin{IEEEbiography}[{\includegraphics[width=1in,height=1.25in,clip,keepaspectratio]{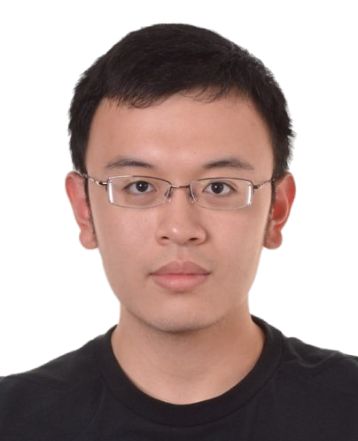}}] {Yunfan Liu} received the B.Eng. degree in electronic engineering from Tsinghua University, China in 2015, the M.S. degree in electronic engineering systems from University of Michigan, USA in 2017, and the Ph.D. degree in Applied Computer Technology from University of Chinese Academy of Sciences, China in 2023. He is a postdoctoral fellow with the School of Electronic, Electrical and Communication Engineering, University of Chinese Academy of Sciences, China. 
His research interests include computer vision and generative models.
\end{IEEEbiography}

\begin{IEEEbiography}[{\includegraphics[width=1in,height=1.25in,clip,keepaspectratio]{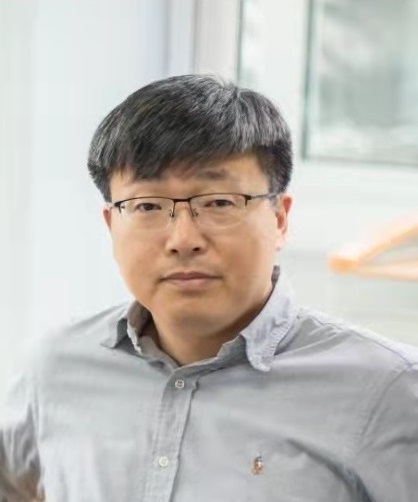}}] {Qixiang Ye} (M'10-SM'15) received the B.S. and M.S. degrees in mechanical and electrical engineering from Harbin Institute of Technology, China, in 1999 and 2001, respectively, and the Ph.D. degree from the Institute of Computing Technology, Chinese Academy of Sciences in 2006. He has been a professor with the University of Chinese Academy of Sciences since 2009, and was a visiting assistant professor with the Institute of Advanced Computer Studies (UMIACS), University of Maryland, College Park until 2013. His research interests include image processing, object detection, and machine learning. He has published more than 100 papers in refereed conferences and journals including IEEE CVPR, ICCV, ECCV, NeurIPS, TPAMI, TIP and TNNLS. 
He was on the editorial board of IEEE Transactions on Circuit and Systems on Video Technology and IEEE Transactions on Intelligent Transportation Systems. 
\end{IEEEbiography}

\vfill

\end{document}